\documentclass{article}

\usepackage{microtype}
\usepackage{graphicx}
\usepackage{subfigure}
\usepackage{booktabs} 
\usepackage{times}
\usepackage{epsfig}
\usepackage{amsmath}
\usepackage{amssymb}
\usepackage{pifont}

\usepackage{multirow}
\usepackage{array}
\newcolumntype{C}[1]{>{\centering\arraybackslash}p{#1}}
\usepackage{hyperref}

\usepackage[accepted]{icml2021}

\icmltitlerunning{Zero-Shot Knowledge Distillation from a Decision-Based Black-Box Model}

\begin{document}

\twocolumn[
\icmltitle{Zero-Shot Knowledge Distillation from a Decision-Based Black-Box Model}



\icmlsetsymbol{equal}{*}

\begin{icmlauthorlist}
\icmlauthor{Zi Wang}{to}
\end{icmlauthorlist}

\icmlaffiliation{to}{Department of Electrical Engineering and Computer Science, The University of Tennessee, Knoxville, TN, USA}

\icmlcorrespondingauthor{Zi Wang}{zwang84@vols.utk.edu}

\icmlkeywords{Machine Learning, ICML}

\vskip 0.3in
]



\printAffiliationsAndNotice{}  

\begin{abstract}
Knowledge distillation (KD) is a successful approach for deep neural network acceleration, with which a compact network (student) is trained by mimicking the softmax output of a pre-trained high-capacity network (teacher). In tradition, KD usually relies on access to the training samples and the parameters of the white-box teacher to acquire the transferred knowledge. However, these prerequisites are not always realistic due to storage costs or privacy issues in real-world applications. Here we propose the concept of decision-based black-box (DB3) knowledge distillation, with which the student is trained by distilling the knowledge from a black-box teacher (parameters are not accessible) that only returns classes rather than softmax outputs. We start with the scenario when the training set is accessible. We represent a sample's robustness against other classes by computing its distances to the teacher's decision boundaries and use it to construct the soft label for each training sample. After that, the student can be trained via standard KD. We then extend this approach to a more challenging scenario in which even accessing the training data is not feasible. We propose to generate pseudo samples distinguished by the teacher's decision boundaries to the largest extent and construct soft labels for them, which are used as the transfer set. We evaluate our approaches on various benchmark networks and datasets and experiment results demonstrate their effectiveness. Codes are available at: \emph{https://github.com/zwang84/zsdb3kd}.
\end{abstract}

\section{Introduction}
Training compact deep neural networks (DNNs) \cite{howard2017mobilenets} efficiently has become an appealing topic because of the increasing demand for deploying DNNs on resource-limited devices such as mobile phones and drones \cite{moskalenko2018model}. Recently, a large number of approaches have been proposed for training lightweight DNNs with the help of a cumbersome, over-parameterized model, such as network pruning \cite{li2016pruning,he2019filter,wang2021convolutional}, quantization \cite{han2015deep}, factorization \cite{jaderberg2014speeding}, and knowledge distillation (KD) \cite{hinton2015distilling,phuong2019towards,jin2020uncertainty,yun2020regularizing,passalis2020heterogeneous,wang2021data}. Among all these approaches, knowledge distillation is a popular scheme with which a compact student network is trained by mimicking the softmax output (class probabilities) of a pre-trained deeper and wider teacher model \cite{hinton2015distilling}. By doing so, the rich information learned by the powerful teacher can be imitated by the student, which often exhibits better performance than solely training the student with a cross-entropy loss. Many variants have been developed to improve the vanilla KD approach by not only mimicking the softmax output but also matching extra elements in the teacher.

The success of KD relies on three factors: (1) access to the teacher's training dataset, (2) the white-box teacher model, i.e., access to the teacher's parameters, and (3) the score-based outputs, i.e., class probabilities of the training samples outputted by the teacher. In real-world applications, however, these prerequisites are usually unrealistic. Due to storage costs of large training datasets (such as ImageNet \cite{deng2009imagenet}) or privacy issues (such as sensitive patient data or personal photos), accessing the training samples are sometimes not feasible. With this concern, the concept of zero-shot knowledge distillation (ZSKD) \cite{nayak2019zero,chen2019data,yin2020dreaming,wang2021data} is proposed.  ZSKD generates pseudo training samples via backpropagation with access to the parameters of the white-box teacher, which are then used as the transfer set for training the student model via KD. However, we argue that this scenario is still not realistic under certain circumstances.

In some cases, training samples are publicly available, but pre-trained models are not. For example, YouTube's recommendation system \cite{covington2016deep} is trained with tons of videos that can be accessed by any user. However, the trained model is a core competitiveness of the company and its parameters are not released. One can argue that a surrogate teacher can be trained locally with the accessible training set, but due to the limitations such as computing resources, its performance is usually not satisfactory compared to the provided powerful model with much more parameters and complicated architectures. 

Moreover, a much more challenging scenario is that, in many real-world applications, none of the three factors mentioned above is available. A pre-trained model stored on the remote server may only provide APIs for inference, neither the model parameters nor the training samples are accessible to the users. Worse than that, these APIs usually return a category index for each sample (i.e., hard-label), rather than the class probabilities over all classes. For example, speech recognition systems like Siri and Cortana are trained with internal datasets and only return the results to users \cite{lopez2017alexa}. Cloud-based object classification systems like Clarifai \cite{clarifai} just give the top-1 classes of the identified objects in the images uploaded by users.

With these concerns, we propose the concept of decision-based black-box knowledge distillation (DB3KD), i.e., training a student model by transferring the knowledge from a black-box teacher that only returns hard-labels rather than probability distributions. We start with the scenario when the training data is available. Our key idea is to extract the class probabilities of the training samples from the DB3 teacher. We claim that the decision boundary of a well-trained model distinguishes the training samples of different classes to the largest extent. Therefore, the distance from a sample to the targeted decision boundary (the boundary to the samples of a certain class) can be used as a representation of a sample's robustness, which determines how much confidence of a specific class is assigned to the sample. Based on this, the soft label of each training sample can be constructed with the value of sample robustness and used for training the student via KD.

We further extend DB3KD to the scenario when training data are not accessible. As the decision boundary makes every effort to differentiate the training samples of all classes, samples used for training the teacher tend to be with longer distances to the boundary than others. We propose to optimize randomly generated noises away from the boundary to obtain robust pseudo samples that simulate the distribution of the training samples. This is achieved by iteratively estimating the gradient direction on the boundary and pushing the samples away from the boundary in that direction. After that, pseudo samples are used for training the student via DB3KD. To our best knowledge, this is the first study of KD from a DB3 teacher, both with and without access to the training set.

The contribution of this study is summarized as follows. (1) We propose the concept of decision-based black-box knowledge distillation for the first time, with which a student is trained by transferring knowledge from a black-box teacher that only returns hard-labels. (2) We propose to use sample robustness, i.e., the distance from a training sample to the decision boundaries of a DB3 teacher, to construct soft labels for DB3KD when training data is available. (3) We extend the DB3KD approach to a more challenging scenario when accessing training data is not feasible and name it zero-shot decision-based black-box knowledge distillation (ZSDB3KD). 
(4) Extensive experiments validate that the proposed approaches achieve competitive performance compared to existing KD methods in more relaxed scenarios.


\section{Related Work}
{\noindent \bf Knowledge distillation.}
Knowledge distillation is first introduced in \cite{bucilua2006model} and generalized in \cite{ba2014deep,hinton2015distilling}, which is a popular network compression scheme to train a compact student network by mimicking the softmax output predicted by a high-capacity teacher or ensemble of models. Besides transferring the knowledge of class probabilities, many variants have been proposed to add extra regulations or alignments between the teacher and the student to improve the performance \cite{romero2014fitnets,yim2017gift,kim2018paraphrasing,heo2019knowledge}. For example, FitNet \cite{romero2014fitnets} introduces an extra loss term that matches the values of the intermediate hidden layers of the teacher and the student, which allows fast training of deeper student models. \cite{zagoruyko2016paying} defines the attention of DNNs and uses it as the additional transferred knowledge. 

{\noindent \bf Knowledge distillation with limited data.} To mitigate the storage and transmission costs of large training datasets, several studies propose the concept of few-shot KD, which generates pseudo samples with the help of a small number of the original training samples \cite{kimura2018few,wang2020neural,li2020few}. 
Another study suggests that instead of the raw data, some surrogates with much smaller sizes (also known as metadata) can be used to distill the knowledge from the teacher. \cite{lopes2017data} leverages the statistical features of the activations of the teacher to train a compact student without access to the original data. However, releasing this kind of metadata along with the pre-trained teacher is usually not a common scenario.

{\noindent \bf Zero-shot knowledge distillation.}
To deal with the scenario when training data is not accessible, \cite{nayak2019zero} proposes zero-shot knowledge distillation (ZSKD). The authors model the softmax output space of the teacher with a Dirichlet distribution and samples soft labels as the targets. Randomly generated noise inputs are optimized towards these targets via backpropagation and are used as the transfer set. \cite{wang2021data} replaces the Dirichlet distribution with a multivariate normal distribution to model the softmax output space of the generated samples. Therefore, pseudo samples of different classes can be generated simultaneously rather than one after another as in \cite{nayak2019zero}. Generative adversarial networks (GANs) \cite{goodfellow2014generative} are leveraged in \cite{chen2019data, micaelli2019zero} to solve this task so that pseudo sample synthesis and student network training can be conducted simultaneously. Another study \cite{yin2020dreaming} proposes to use the features in the batch normalization layers to generate pseudo samples. 
However, these methods still need access to the parameters of the teacher for backpropagation, which is unrealistic in many cases.

{\noindent \bf Black-box knowledge distillation.}
Although the vanilla KD is built with a black-box teacher \cite{hinton2015distilling}, the whole training dataset is used for training. \cite{wang2020neural} investigates the possibility that a student is trained with limited samples and a black-box teacher. Other than zero-shot KD methods that generate pseudo inputs, \cite{orekondy2019knockoff} proposes to sample from a large pool (such as ImageNet) to get the transfer set to train the student. Therefore, there is no need to access the teacher's parameters.
Although the prerequisites in these methods are relaxed, weak assumptions on the training samples and a score-based teacher that outputs class probabilities are still needed. Different from these studies, we consider a much more challenging case in which knowledge is transferred from a black-box teacher that only returns top-1 classes.


{\noindent \bf Decision-based adversarial attack.} Our approach leverages the distance from a sample to the decision boundary for soft label construction, which is related to the research of decision-based black-box adversarial attack \cite{brendel2017decision,cheng2018query,cheng2019sign,liu2019geometry}. These methods aim to add some imperceptible perturbations to the inputs to create adversarial samples that fool a well-trained DNN with high confidence. This is achieved by identifying the points on the decision boundary with minimal distance to the original inputs. Inspired by these studies, we use the distance from a sample to the targeted decision boundaries as a representation of a sample's robustness against other categories, which can be converted to a probability distribution of all classes with proper operations.


\begin{figure*}[t]
\begin{center}
\includegraphics[width=0.98\linewidth]{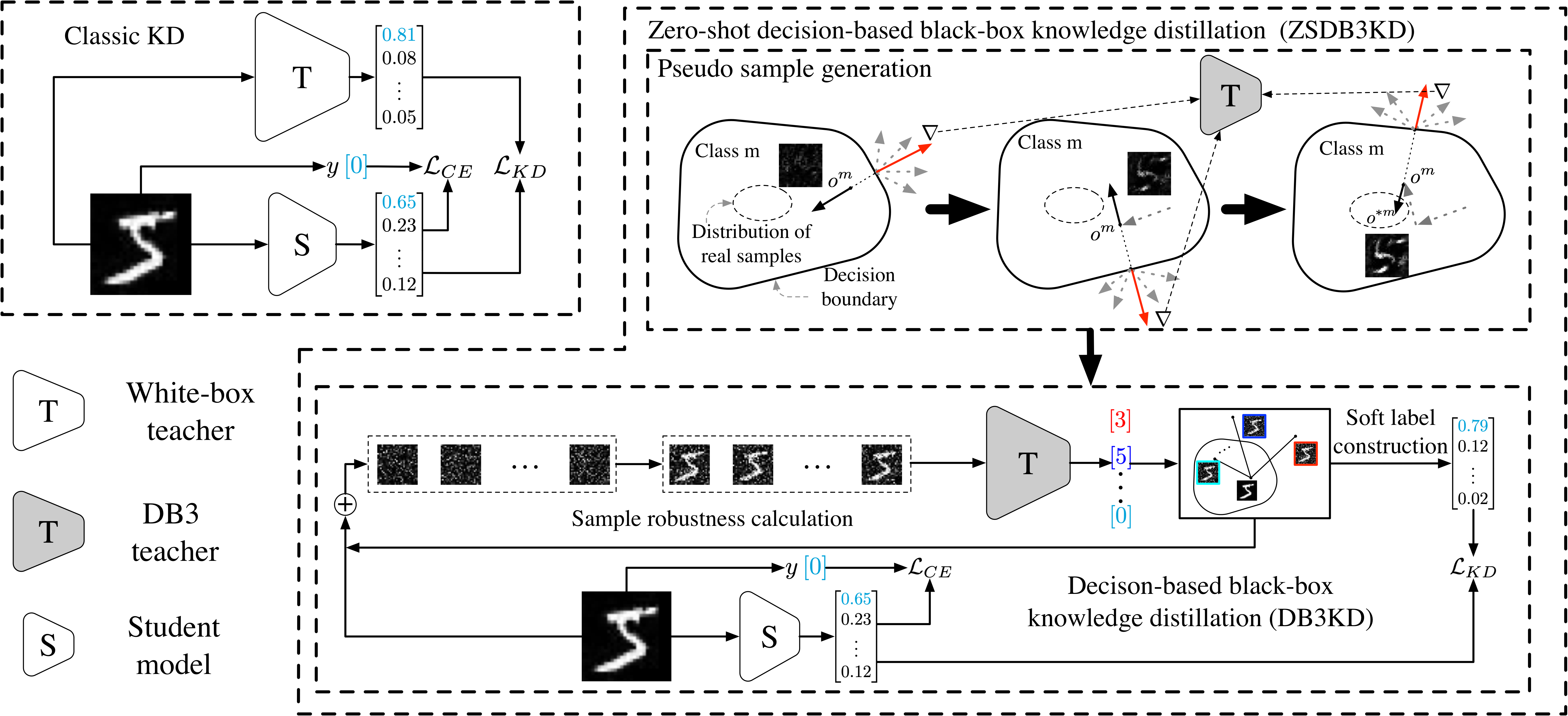}
\end{center}
\vskip -0.15in
\caption{The overall workflow of the proposed approach. Left: classic KD. Bottom: decision-based black-box KD (DB3KD). Samples are iteratively fed to the DB3 teacher to compute the sample robustness, which is transformed as soft labels for training the student via KD. Right: Zero-shot DB3KD (ZSDB3KD). Pseudo samples are generated by moving random noises away from the decision boundary and approaching the distribution of the original training samples, which are used as the transfer set for training the student via DB3KD.}
\label{fig:whole_pic}
\vskip -0.2in
\end{figure*}

\section{Methodology}
We first formulate KD in its standard form and present our approach that creates soft labels of the training samples with a DB3 teacher. Finally, we extend our approach to the scenario in which the training set is not accessible.

\subsection{Knowledge Distillation}
KD is used for training a compact student by matching the softmax outputs of a pre-trained, cumbersome teacher \cite{hinton2015distilling} (Fig.~\ref{fig:whole_pic}(left)). For an object classification task, denote $F_t(x)$ and $F_s(x)$ the teacher and the student DNNs, respectively, which take an image $x$ as the input, and output a vector $P \in \left[0,1\right]^L$, i.e., $F_t(x)=P_t=\text{softmax}(a_t)$, $F_s(x)=P_s=\text{softmax}(a_s)$, where $L$ is the number of classes and $a$ is the pre-softmax activation. In a KD procedure, a temperature $\tau$ is usually introduced to soften the softmax output, i.e., $P^\tau=\text{softmax}(a/\tau)$, which is proved to be efficient to boost the training process. 
The student is trained by minimizing the loss function in Eq. \eqref{eq:classic_kd}.
\begin{equation}
\mathcal{L}=\mathcal{L}_{CE}(P_s,y)+\lambda \mathcal{L}_{KD}(P^\tau_t,P^\tau_s),
\label{eq:classic_kd}
\end{equation}
where $y$ is the ground truth label, $\mathcal{L}_{CE}$ and $\mathcal{L}_{KD}$ are the cross-entropy loss and the distillation loss. A scaling factor $\lambda$ is used for balancing the importance of the two losses.

\begin{figure}[t]
\begin{center}
\includegraphics[width=0.95\linewidth]{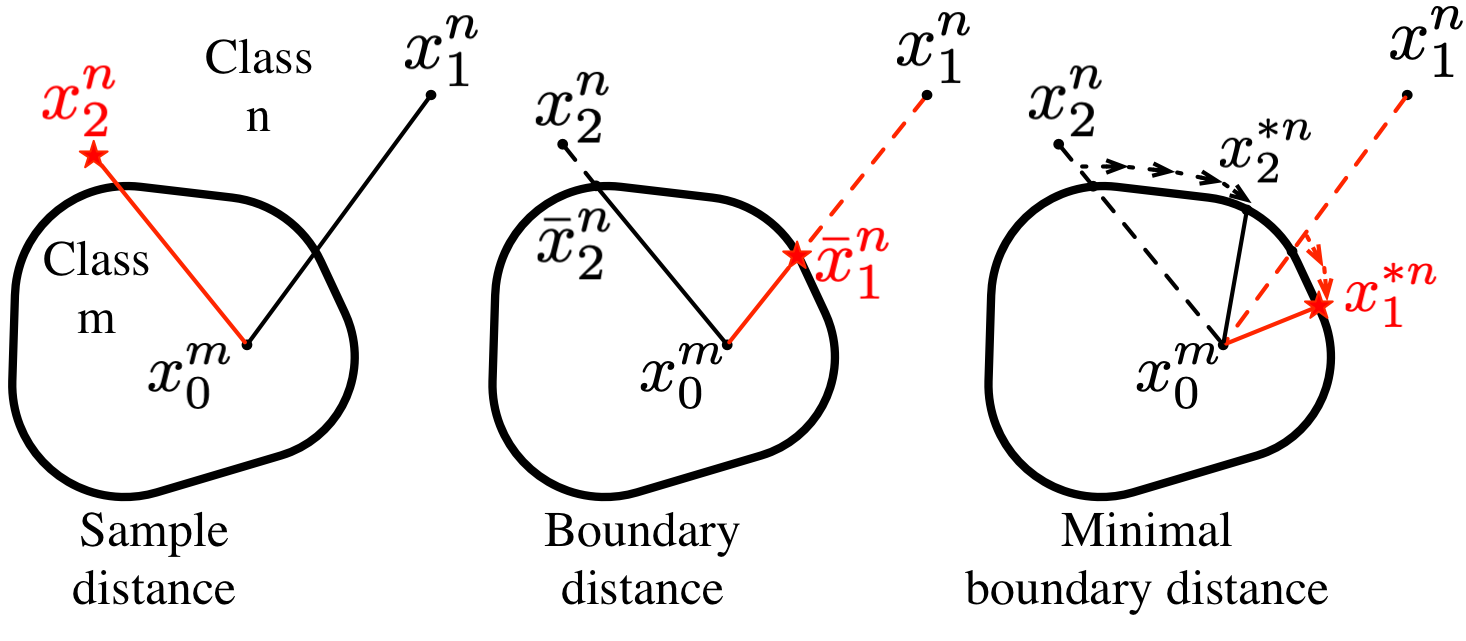}
\end{center}
\vskip -0.2in
\caption{Strategies for computing sample robustness.}
\label{fig:long}
\vskip -0.2in
\end{figure}
\subsection{Decision-Based Black-Box Knowledge Distillation}
\label{sec:db3}
As mentioned, in many real-world applications, users are prohibited from querying any internal configuration of the teacher except for the final decision (top-1 label). Denote $F_t^{B}(x)$ the DB3 teacher, then $F_t^{B}(x)=l, l \in \{1,2,\cdots,L\}$. In this case, $P_t$ cannot be obtained and the student cannot be trained with Eq. \eqref{eq:classic_kd}. We claim that a sample's robustness against a specific class can be used as a representation of how much confidence should be assigned to this class, with proper post-operations. Therefore, we extract the sample's robustness against each class from the DB3 teacher and convert it to a class distribution $\hat{P_t}$ as an estimate of $P_t$ (Fig.~\ref{fig:whole_pic}(bottom)). In the following, we propose three metrics to measure sample robustness and present how to construct class distributions with the sample robustness measurements. Intuitively, if a sample is closer to some points in the region of a specific class, it is more vulnerable to this class and thus should be assigned higher confidence.

\subsubsection{Sample Robustness}
\label{sec:s_vul}
{\noindent \bf Sample Distance (SD).} The most straightforward way to quantify the sample robustness is to compute the minimal $\ell_2$-norm distance from a sample to those of other classes (Fig.~\ref{fig:long}(left)). Denote $x_0^{m} \in \mathbb{R}^{C \times W \times H}$ a sample of the $m$-th class, $\mathbf{x}^{n}=\{x_1^{n},x_2^{n},\cdots,x_S^{n}\}$ a batch of $S$ samples from the $n$-th class, where $n \neq m$, $C,W,H$ are the number of channels, width and height of the sample, respectively. The robustness of $x_0^{m}$ against class $n$ is computed with Eq.~\eqref{eq:sample_dist}.
\begin{equation}
r_0^{m,n} = \min_{1 \le i \le S} ||x_i^{n} - x_0^{m}||_2.
\label{eq:sample_dist}    
\end{equation}
The advantage of using SD is it can be implemented without querying from the teacher. However, SD is a rough estimate of sample robustness since it does not mine any information from the teacher. Therefore, we introduce two advanced strategies to measure sample robustness.

{\noindent \bf Boundary Distance (BD).} To obtain better representation of sample robustness, we propose to leverage the distances from a sample to the targeted decision boundaries of the teacher (Fig.~\ref{fig:long}(middle)). For each $x_i^{n} \in \mathbf{x}^{n}$, we implement a binary search in the direction $(x_i^{n}-x_0^{m})$ and find the corresponding point $\bar{x}_i^{n}$ on the decision boundary (Eq.~\eqref{eq:sample_on_boundary}).
\vskip -0.2in
\begin{equation}
\begin{aligned}
\bar{x}_i^{n} = \min_\alpha (&x_0^{m} + \alpha \cdot \frac{x_i^{n}-x_0^{m}}{||x_i^{n}-x_0^{m}||_2}), i=1,2,\cdots,S, \\
&\text{s.t.}~~F_t^B(\bar{x}_i^{n}+\epsilon) = n, ~~~~||\epsilon||_2 \to 0.
\end{aligned}
\label{eq:sample_on_boundary}    
\end{equation}
\vskip -0.1in
We then compute the sample robustness with Eq.~\eqref{eq:sample_dist} in which $x_i^{n}$ is replaced by $\bar{x}_i^{n}$.

\begin{figure}[t]
\begin{center}
\includegraphics[width=0.8\linewidth]{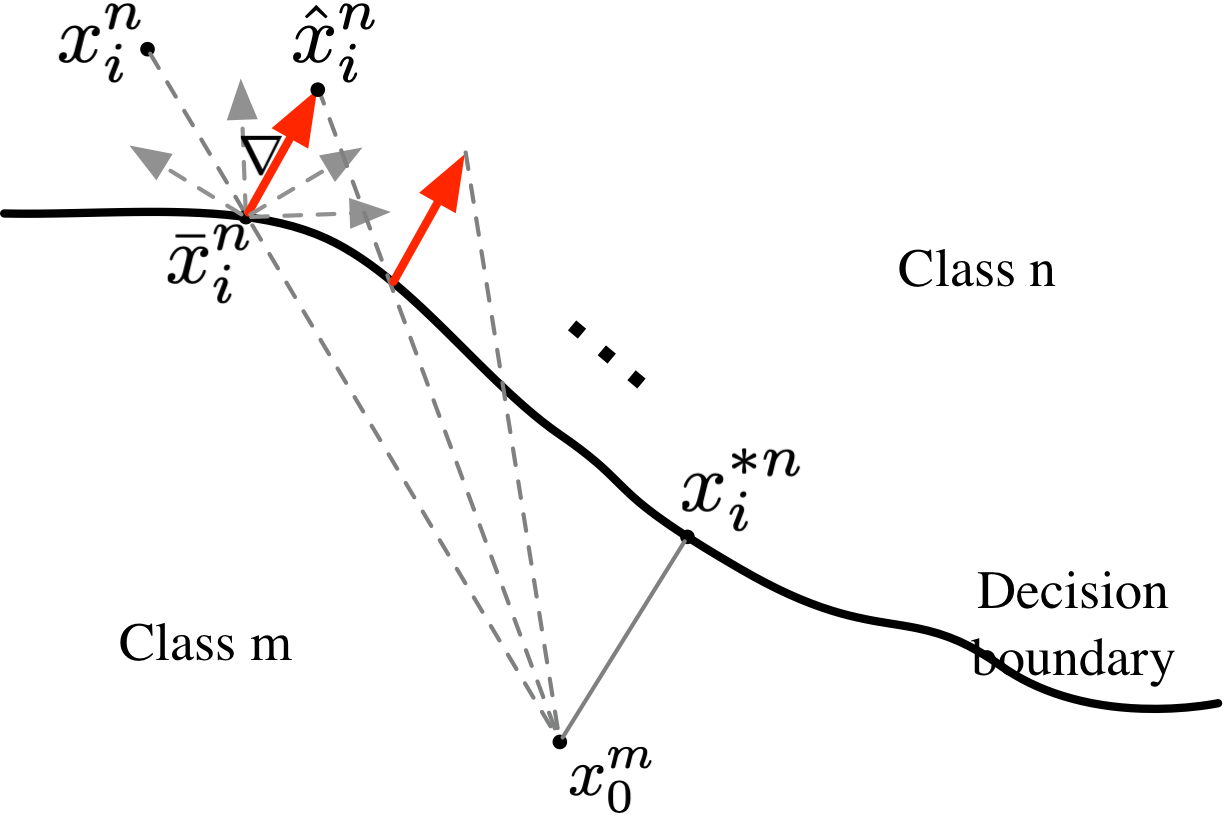}
\end{center}
\vskip -0.15in
\caption{The iterative procedure for the optimization of MBD.}
\label{fig:mbd}
\vskip -0.15in
\end{figure}

{\noindent \bf Minimal Boundary Distance (MBD).}
Inspired by recent studies of decision-based black-box adversarial attack \cite{brendel2017decision,cheng2018query,liu2019geometry,cheng2019sign}, we further optimize $\bar{x}_i^{n}$ by moving it along the decision boundary to the point $x_i ^{*n}$ where $||x_i ^{*n} - x_0^{m}||_2$ is minimized (Fig.~\ref{fig:long}(right)). Starting from $\bar{x}_i^{n}$, we first estimate the gradient of the boundary $\nabla F_t^B(\bar{x}_i^{n})$ via zeroth order optimization \cite{wang2018stochastic}, which is achieved by sampling $Q$ Gaussian random vectors $\mathbf{u}_q \in \mathbb{R}^{C \times W \times H}~ (q=1,2,\cdots,Q)$ and averaging them (Fig.~\ref{fig:mbd}, Eq.~\eqref{eq:gradient_est}).
\begin{equation}
\nabla F_t^B(\bar{x}_i^{n}) = \frac{1}{Q}\sum_{q=1}^Q \text{sign}(\bar{x}_i^{n} + \epsilon_g \mathbf{u}_q) \mathbf{u}_q,
\label{eq:gradient_est}    
\end{equation}
where $\epsilon_g$ is a very small scalar, and $\text{sign}(x_i ^{n} + \epsilon_g \mathbf{u}_q)$ is a sign function, i.e, 
\begin{equation}
\text{sign}(x_i ^{n} + \epsilon_g \mathbf{u}_q)=
\begin{cases}
        +1,~~F_t^B(\bar{x}_i^{n} + \epsilon_g \mathbf{u}_q)=n,\\
        -1,~~\text{Otherwise}.\\
\end{cases}
\label{eq:sign}
\end{equation}
Once the gradient is determined, we get a new sample outside the decision boundary $\hat{x}_i^{n} \leftarrow \bar{x}_i^{n}+\xi_d \nabla F_t^B(\bar{x}_i^{n})$ with a step size $\xi_d$. Then we conduct the same binary search procedure (Eq.~\eqref{eq:sample_on_boundary}) in the direction $(\hat{x}_i^{n}-x_0^{m})$ and obtain an updated $\bar{x}_i^{n}$. Since the search is within a very small region, the decision boundary in such a region is smooth. Therefore, the new $\bar{x}_i^{n}$ has a smaller distance to $x_0^{m}$ (Fig.~\ref{fig:mbd}). We repeat the procedure above to get the optimal solution $x_i^{*n}=\bar{x}_i^{n}$ until $||\bar{x}_i^{n} - x_0^{m}||_2$ cannot be further minimized or the query limit is reached. Finally, we compute the sample robustness with Eq.~\eqref{eq:sample_dist} in which $x_i^{n}$ is replaced by $x_i^{*n}$.


\subsubsection{Soft Label Construction}
After obtaining all the samples' robustness on all classes, we construct the soft labels for them with proper manipulations. We start with the pre-softmax activations for better illustration. Suppose the pre-softmax activation of a sample $x_s^{m}$ is $\mathbf{a_s^{m}}=\{a_{s,1}^{m},a_{s,2}^{m},\cdots,a_{s,L}^{m}\}$. Then the pre-softmax activation and the sample robustness should be in correlation with the following conditions. (1) $\text{argmax}_i a_{s,i}^{m}=m$. It is obvious that $a_{s,m}^{m}$ should be the largest number to ensure that the sample is assigned to the correct class. (2) If $r_s^{m,j} > r_s^{m,k}$, then $a_{s,j}^{m} < a_{s,k}^{m}$. This is because bigger sample robustness indicates longer distance to the targeted decision boundary, which means that the sample is more robust against the certain class and should be assigned a lower confidence. (3) If $\sum_{j=1}^L r_s^{m,j} > \sum_{j=1}^L r_p^{m,j}, j \neq m$, then $a_{s,m}^{m} > a_{p,m}^{m}$. This is because when the sum of a sample's distances to its targeted decision boundaries is larger, the probability mass of this sample is more concentrated in its top-1 class. Otherwise, the mass is more dispersed among all elements.






With the above design philosophy, to meet requirement (1) and (2), we define $\hat{a}_{s,n}^{m} (n=1,2,\cdots,L)$ in Eq.~\eqref{eq:logits}.
\begin{equation}
\hat{a}_{s,n}^{m}=\begin{cases}
        \frac{1}{r_s^{m,n}},~~~~\text{for}~n \neq m, \\
        \sum_{i=1}^L \frac{1}{r_s^{m,i}}, i \neq m,~~~~\text{for}~n = m.\\
    \end{cases}
\label{eq:logits}
\end{equation}
$\hat{a}_{s,n}^{m}$ is then divided by $(\sum_{i=1}^L \frac{1}{r_s^{m,i}})^2$ to meet requirement (3), as presented in Eq.~\eqref{eq:logits_final}.
\begin{equation}
{a}_{s,n}^{m}=\frac{\hat{a}_{s,n}^{m}}{(\sum_{i=1}^L \frac{1}{r_s^{m,i}})^2},~~i \neq m,~\text{for}~n=1,2,\cdots,L.
\label{eq:logits_final}    
\end{equation}
Finally, we get $\hat{P}_t=\text{softmax}(\mathbf{a}_{s}^{m})$ for sample $x_s^{m}$.

\subsubsection{Training of Student Model}
Once the soft labels of all the training samples are constructed with the above approach, we can train the student with standard KD, using the objective function in Eq.~\eqref{eq:classic_kd}.

\subsection{Zero-shot Decision-Based Black-Box Knowledge Distillation}
In zero-shot KD, pseudo samples are usually generated by optimizing some noise inputs via backpropagation towards some soft labels sampled from a prior distribution, which are then used as the transfer set. However, with a DB3 teacher, backpropagation cannot be implemented and the prior distribution cannot be obtained, which makes ZSDB3KD a much more challenging task. Since the teacher is trained to largely distinguish the training samples, the distance between a training sample to the teacher's decision boundary is usually much larger than the distance between a randomly generated noise image to the boundary. With this claim, we propose to iteratively push random noise inputs towards the region that is away from the boundary to simulate the distribution of the original training data (Fig.~\ref{fig:whole_pic}(right)).

Denote $o_0^{m}$ and $\mathbf{o}^{\bar{m}}=\left[o_1^{\bar{m}},o_2^{\bar{m}},\cdots,o_T^{\bar{m}}\right]$ a random noise input of the $m$-th class and a batch of $T$ random noises with any other class, respectively. Similar but slightly different from Eq.~\eqref{eq:sample_on_boundary}, for each $o_i^{\bar{m}} \in \mathbf{o}^{\bar{m}}$, we first identity its corresponding points on the boundary $\bar{o}_i^{m}$ with Eq.~\eqref{eq:untargeted_BD}.
\begin{equation}
\begin{aligned}
\bar{o}_i^{m} = \min_\alpha (&o_0^{m} + \alpha \cdot \frac{o_i^{\bar{m}}-o_0^{m}}{||o_i^{\bar{m}}-o_0^{m}||_2}), i=1,2,\cdots,T, \\
&\text{s.t.}~~F_t^B(\bar{o}_i^{m}+\epsilon) \neq m, ~~~~||\epsilon||_2 \to 0.
\end{aligned}
\label{eq:untargeted_BD}
\end{equation}
Similarly, the MBDs of $o_0^{m}$, i.e., $o_i^{*m}$, can be iteratively estimated with Eq.~\eqref{eq:gradient_est} and \eqref{eq:sign}. Let  $o^{*m}$ be the one of $o_i^{*m}~(i=1,2,\cdots,T)$ such that $||o^{*m} - o_0^{m}||_2$ attains its minimal value, i.e., $||o^{*m} - o_0^{m}||_2=\min_i ||o_i^{*m} - o_0^{m}||_2$.
We then estimate the gradient at the boundary $\nabla F_t^B({o}^{*m})$ with Eq.~\eqref{eq:gradient_est} and update $o^{m}$ as $o^{m}\leftarrow o^{m}-\xi_o \nabla F_t^B({o}^{*m})$ with the step size $\xi_o$. The new $o^{m}$ is usually with longer distance to the boundary. We repeat the above process until $||o^{*m}-o^{m}||_2$ cannot be further maximized or the query limit is reached. Finally, we used the generated pseudo samples with the DB3KD approach to train the student as described in Section \ref{sec:db3}. 


\section{Experiments}
In this section, we first demonstrate the performance of DB3KD when training samples are accessible. Then we show the results of ZSDB3KD under the circumstance that training data is not accessible.

\begin{table*}[t]
\begin{center}
\begin{tabular}{|c|c|c|c|c|c|c|c|}
\hline
\multirow{3}{*}{Algorithm} & \multicolumn{2}{|c|}{MNIST} & \multicolumn{2}{|c|}{Fashion-MNIST} & \multicolumn{2}{|c|}{CIFAR10} & FLOWERS102 \\
\cline{2-8} 
 & LeNet5 & LeNet5 & LeNet5 & LeNet5 & AlexNet & AlexNet & \multirow{2}{*}{ResNet-18}\\
 & -half & -1/5 & -half & -1/5 & -half & -quarter & \\
\hline\hline
Teacher CE & 99.33\% & 99.33\% & 91.63\% & 91.63\% & 79.30\% & 79.30\% & 95.07\% \\
Student CE  & 99.11\% & 98.77\% & 90.21\% & 88.75\% & 77.28\% & 72.21\% & 92.18\% \\
Standard KD & 99.33\% & 99.12\% & 90.82\% & 89.09\% & 77.81\% & 73.14\% & 94.05\% \\
Surrogate KD & 99.13\% & 98.85\% & 90.27\% & 88.72\% & 77.49\% & 72.49\% & 92.93\% \\
Noise logits & 99.01\% & 98.72\% & 89.81\% & 88.20\% & 77.04\% & 72.06\% & 91.99\%\\
{\bf DB3KD-SD} &  99.15\% & 98.98\% & 90.86\% & 89.31\% & 77.66\% & 72.78\% & 93.18\%\\
{\bf DB3KD-BD} & 99.51\% & 99.19\% & 90.68\% & 89.47\% & 77.92\% & 72.94\% & 93.30\% \\
{\bf DB3KD-MBD} & 99.52\% & 99.22\% & 91.45\% & 89.80\% & 78.30\% & 73.78\% & 93.77\% \\
\hline
\end{tabular}
\end{center}
\vskip -0.1in
\caption{Performance evaluation of the proposed DB3KD approach.}
\label{tab:main}
\vskip -0.1in
\end{table*}

\subsection{Experiment Setup of DB3KD}
\label{sec:setup_DB3}
We demonstrate the effectiveness of DB3KD with several widely used DNNs and datasets as follows. (1) A LeNet-5 \cite{lecun1998gradient} with two convolutional layers is pre-trained on MNIST \cite{lecun1998gradient} as the teacher, following the configurations in \cite{lopes2017data,chen2019data}. A LeNet-5-Half and a LeNet-5-1/5 are designed as the student networks, which contains half and 1/5 number of convolutional filters in each layer compared to LeNet-5, respectively. (2) The same teacher and student networks as in (1) are used but are trained and evaluated on the Fashion-MNIST dataset. (3) An AlexNet \cite{krizhevsky2012imagenet} pre-trained on CIFAR-10 \cite{krizhevsky2009learning} is used as the teacher. An AlexNet-Half and an AlexNet-Quarter with half and 25\% filters are used as student networks. (4) A ResNet-34 \cite{he2016deep} pre-trained on the high-resolution, fine-grained dataset FLOWERS102 \cite{nilsback2008automated} is used as the teacher, and the student is a ResNet-18.

We evaluate our approach with the three strategies for sample robustness calculation as described in Section \ref{sec:s_vul}, represented as DB3KD-SD, DB3KD-BD, and DB3KD-MBD, respectively. For DB3KD-SD, we use 100 samples from each class to compute the sample robustness $r$ for MNIST, Fashion-MNIST, and CIFAR-10. Since there are only 20 samples in each class of FLOWERS102, we use all of them. Starting with these samples, $\epsilon$ is set to $1e^{-5}$ as the stop condition of the binary search in DB3KD-BD. In DB3KD-MBD, we use 200 Gaussian random vectors to estimate the gradient and try different numbers of queries from 1000 to 20000 with $\xi_d=0.2$ to optimize the MBD and report the best test accuracies. The sample robustness are calculated in parallel with a batch size of 20 with FLOWERS102, and 200 with the other datasets.

With the constructed soft labels, we train the student networks for 100 epochs, using an Adam optimizer (learning rate $5e^{-3}$), for all the datasets except for FLOWERS102, which is trained for 200 epochs. The scaling factor $\lambda$ is set to 1 for simplicity. Since Eq.~\eqref{eq:logits_final} has the similar functionality with the temperature $\tau$, $\tau$ is not need to be as large as in previous studies \cite{hinton2015distilling}.With a hyperparameter search, we find that smaller $\tau$s between $0.2$ and $1.0$ leads to good performance. We use $\tau=0.3$ in our experiments. All experiments are evaluated for 5 runs with random seeds.

\begin{table}[t]
\centering
\begin{tabular}{|c|c|c|c|}
\hline
Approach & Teacher & Student & Accuracy \\
\hline
Cross-entropy & ResNet-34 & - & 78.63\% \\
\hline
Cross-entropy & ResNet-18 & - & 75.91\% \\
\hline
Standard KD & \multirow{6}{*}{ResNet-34} & \multirow{6}{*}{ResNet-18} & 77.18\% \\
\cline{1-1}\cline{4-4}
Surrogate KD &  &  & 76.52\% \\
\cline{1-1}\cline{4-4}
BAN$^*$ &  &  & 76.84\% \\
\cline{1-1}\cline{4-4}
TF-KD &  &  & 77.23\% \\
\cline{1-1}\cline{4-4}
SSKD &  &  & 76.20\% \\
\cline{1-1}\cline{4-4}
DB3KD &  &  & 77.31\% \\
\hline
DB3KD & ResNet-50 & ResNet-18 & 78.65\% \\
\hline
\end{tabular}
\caption{Performance comparison to self-distillation approaches with ResNet on CIFAR-100. $*$ indicates the results are based on our own implementation.}
\label{tab:selfkd}
\vskip -0.2in
\end{table}

\begin{figure*}[t]
\begin{center}
\subfigure[LeNet5-MNIST]{
\includegraphics[width=0.24\linewidth]{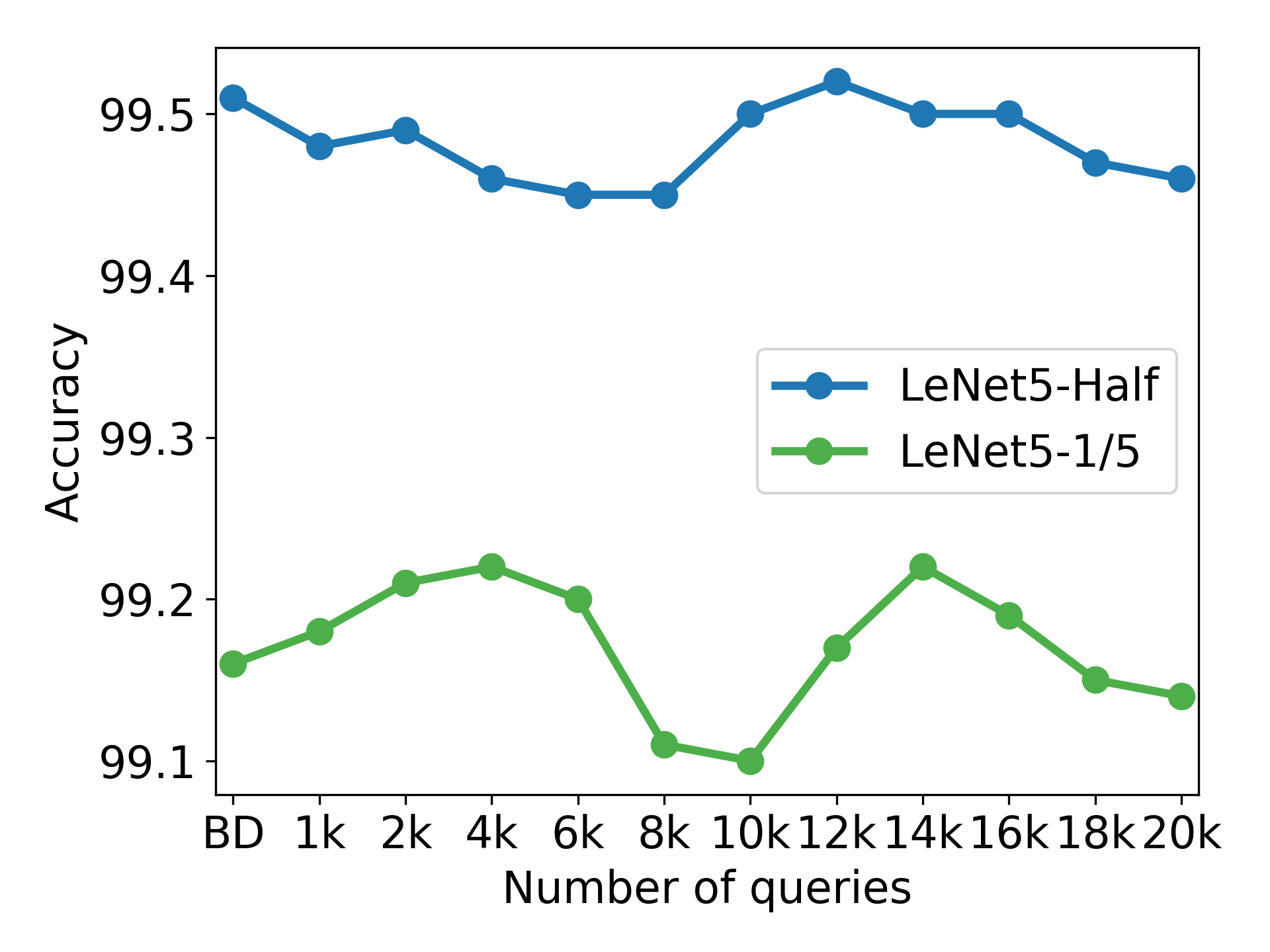}}
\subfigure[LeNet5-Fashion-MNIST]{
\includegraphics[width=0.24\linewidth]{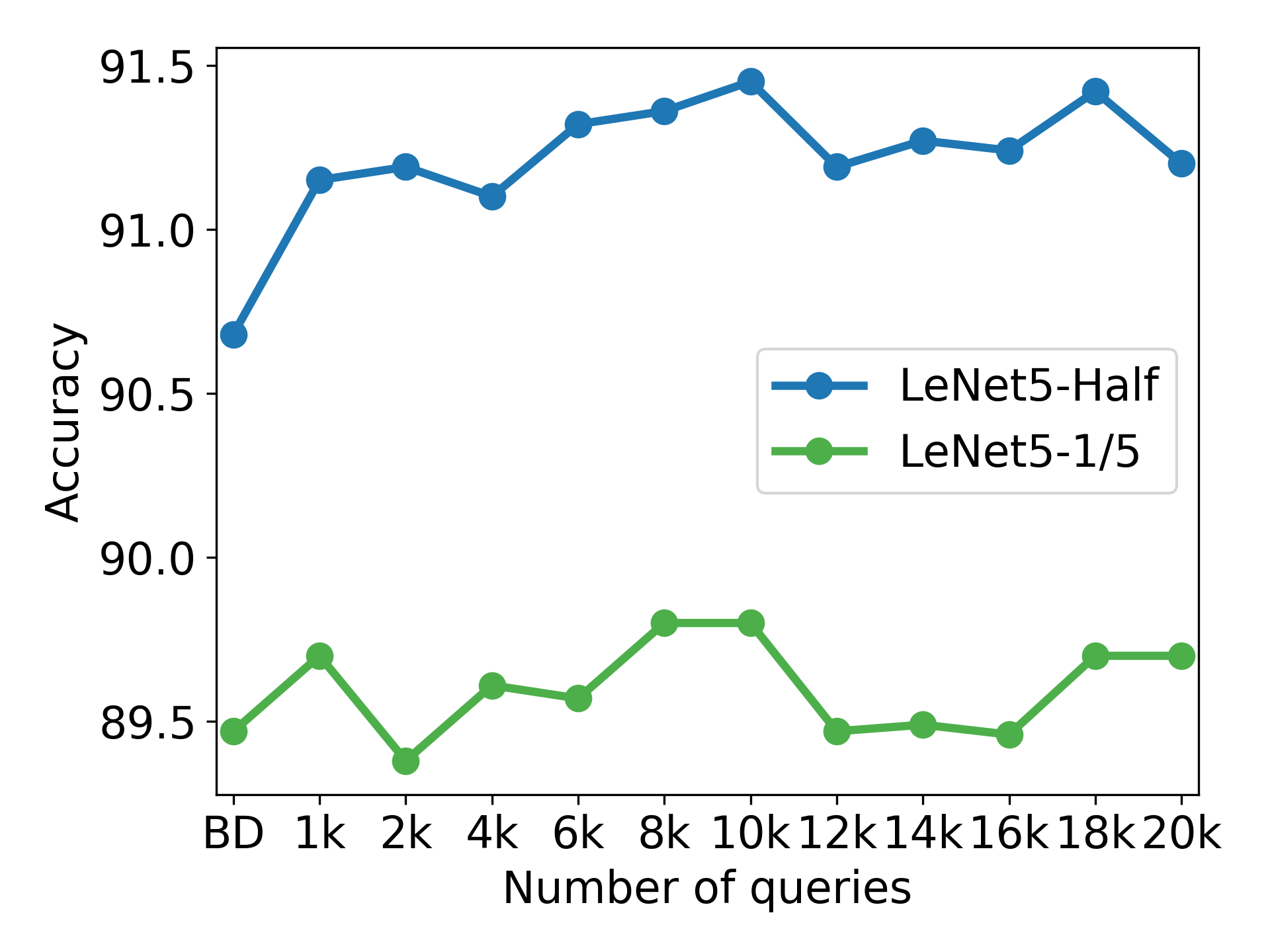}}
\subfigure[AlexNet-CIFAR10]{
\includegraphics[width=0.24\linewidth]{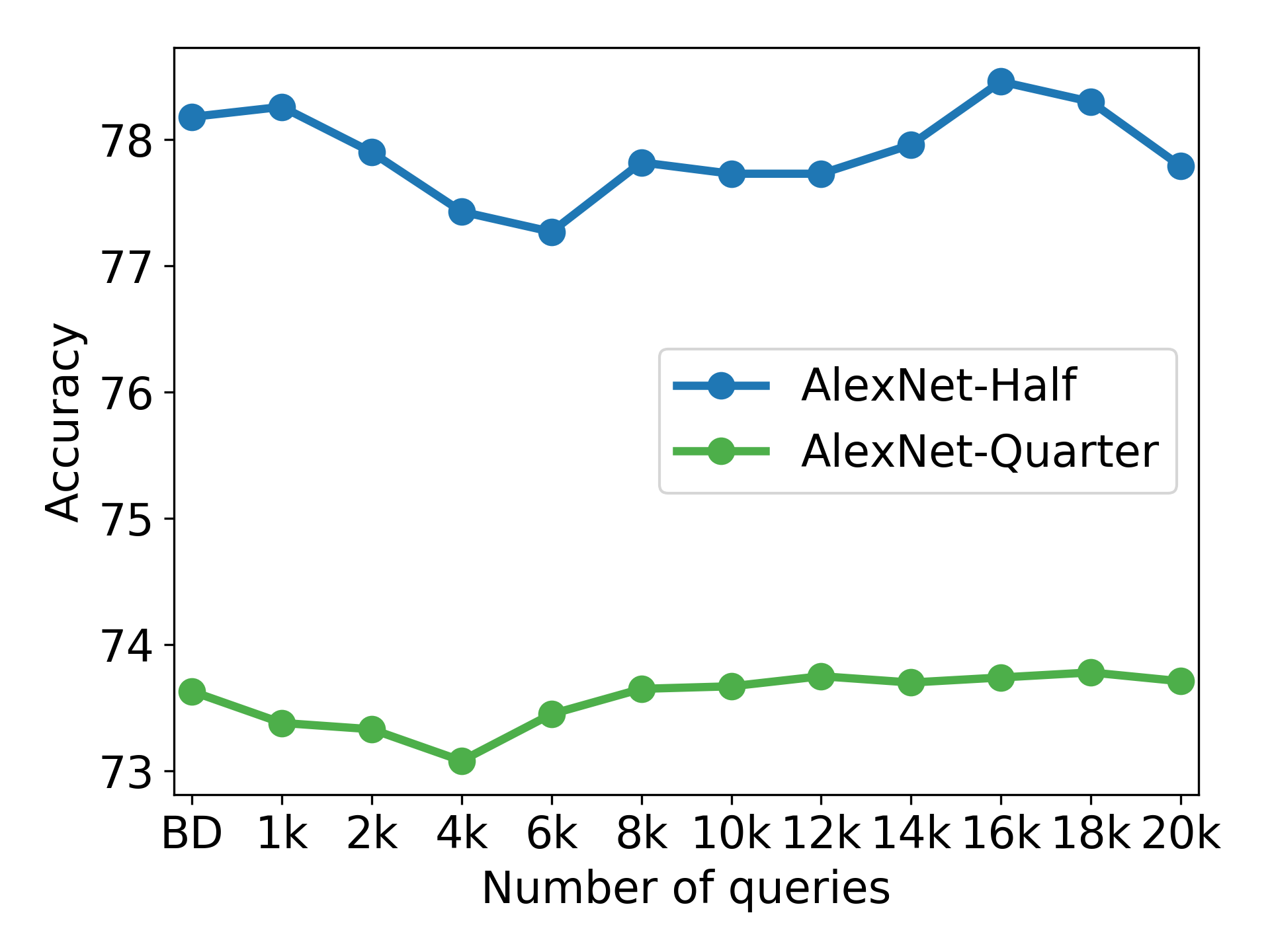}}
\subfigure[ResNet-FLOWERS102]{
\includegraphics[width=0.24\linewidth]{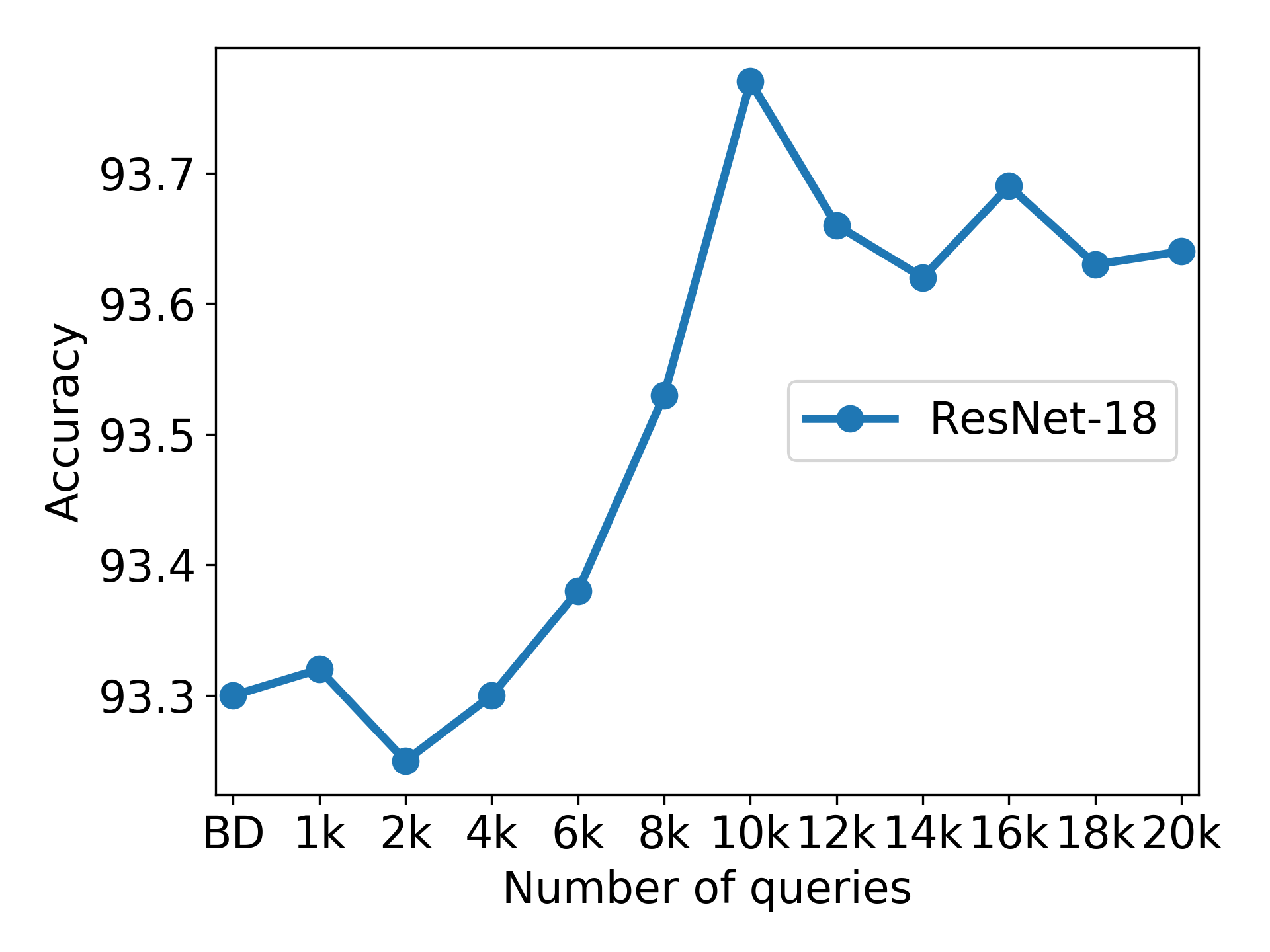}}
\end{center}
\vskip -0.13in
\caption{Performance comparison with different numbers of queries for computing sample robustness.}
\label{fig:iters}
\vskip -0.13in
\end{figure*}

\subsection{Performance Evaluation of DB3KD}
The performance of DB3KD is presented in Table~\ref{tab:main}. To understand the proposed approach better, we also present the performance of the following training strategies. (1) The teacher and the student networks trained solely with the cross-entropy loss. (2) The standard KD with Eq.~\eqref{eq:classic_kd} \cite{hinton2015distilling}. (3) Training the student network via KD with a surrogate white-box teacher (Surrogate KD in Table~\ref{tab:main}), which is used for simulating the scenario in which one can train a smaller but affordable surrogate model with full access to its parameters compared to the powerful DB3 teacher. Here the surrogate has the same architecture with the student. The performance of surrogate KD is considered as the lower bound of DB3KD. (4) Training with the soft labels constructed with randomly generated sample robustness (Noise logits in Table~\ref{tab:main}), which is used for verifying the effectiveness of DB3KD for soft label construction.

We observe from the results that DB3KD works surprisingly well. With the most straightforward strategy SD, our approach still achieve competitive performance on all experiments compared to standard KD and outperform surrogate KD. When using MBD to compute sample robustness, DB3KD-MBD outperforms standard KD on all the experiments except for FLOWERS102. On FLOWERS102, the performance of DB3KD is slightly worse due to the complexity of the pre-trained teacher model. However, DB3KD still outperforms the surrogate KD with a clear margin. These results validate the effectiveness of DB3KD and indicates that sample robustness with proper post-operation provides an informative representation of a sample's probabilities over all classes and can be used as an alternative to the softmax output when only a DB3 teacher is provided.

We also observe the following phenomena in the experiments. (1) Training with noise logits via KD does not work, but even results in worse performance than training with cross-entropy. It indicates noise logits cannot capture the distribution of class probabilities, but are even harmful due to the wrong information introduced. 
(2) Training a student with a surrogate teacher not only results in unsatisfactory performance, but is also a difficult task due to the low capacity of the surrogate model. Also, the performance is sensitive to hyperparameter selection ($\lambda$, $\tau$, learning rate, etc.). Therefore, training an extra affordable surrogate teacher is not an optimal solution compared to DB3KD.

We notice that in some experiments, surprisingly, DB3KD even works better than standard KD, though the models are trained with a more challenging setting. A reasonable hypothesis is that, for some problems, the distance between a training sample to the decision boundary may provide more information than the softmax output. These results provide future research directions that the dark knowledge behind the teacher's decision boundary is more instructive compared to the teacher's logits in certain cases.

\subsection{Comparison with Self-Distillation Approaches}
Similar to our proposed scenario, in the absence of a pre-trained teacher, self-knowledge distillation aims to improve the performance of the student by distilling the knowledge within the network itself \cite{furlanello2018born}. Since self-distillation approaches can also deal with our proposed scenario, we compare the performance of DB3KD to recent self-distillation approaches, including born-again neural networks (BAN) \cite{furlanello2018born}, teacher-free
knowledge distillation (TF-KD) \cite{yuan2020revisiting}, and self-supervision knowledge distillation (SSKD) \cite{xu2020knowledge}. We use ResNet-34/18 as the teacher and the student on CIFAR-100 for illustration. For further comparison, we also implement DB3KD with a ResNet-50 teacher.

The results are shown in Table \ref{tab:selfkd}. It is observed that our approach is still competitive in this case. With the same network configuration, our student achieves a test accuracy of 77.31\%, which outperforms other self-distillation approaches, even with a DB3 teacher. It is also worth mentioning that, given a fixed student, the performance of self-distillation has an upper bound because it is teacher-free. One advantage of our approach is that the student can leverage the information from a stronger teacher and its performance can be further improved. As an example, we substitute the DB3 teacher with a ResNet-50 network and keep other other configurations unchanged, the performance of our student network is further increased by 1.34\%, which outperforms self-distillation approaches with a clear margin.

\begin{figure*}[t]
\begin{center}
\subfigure[]{
\includegraphics[width=0.24\linewidth]{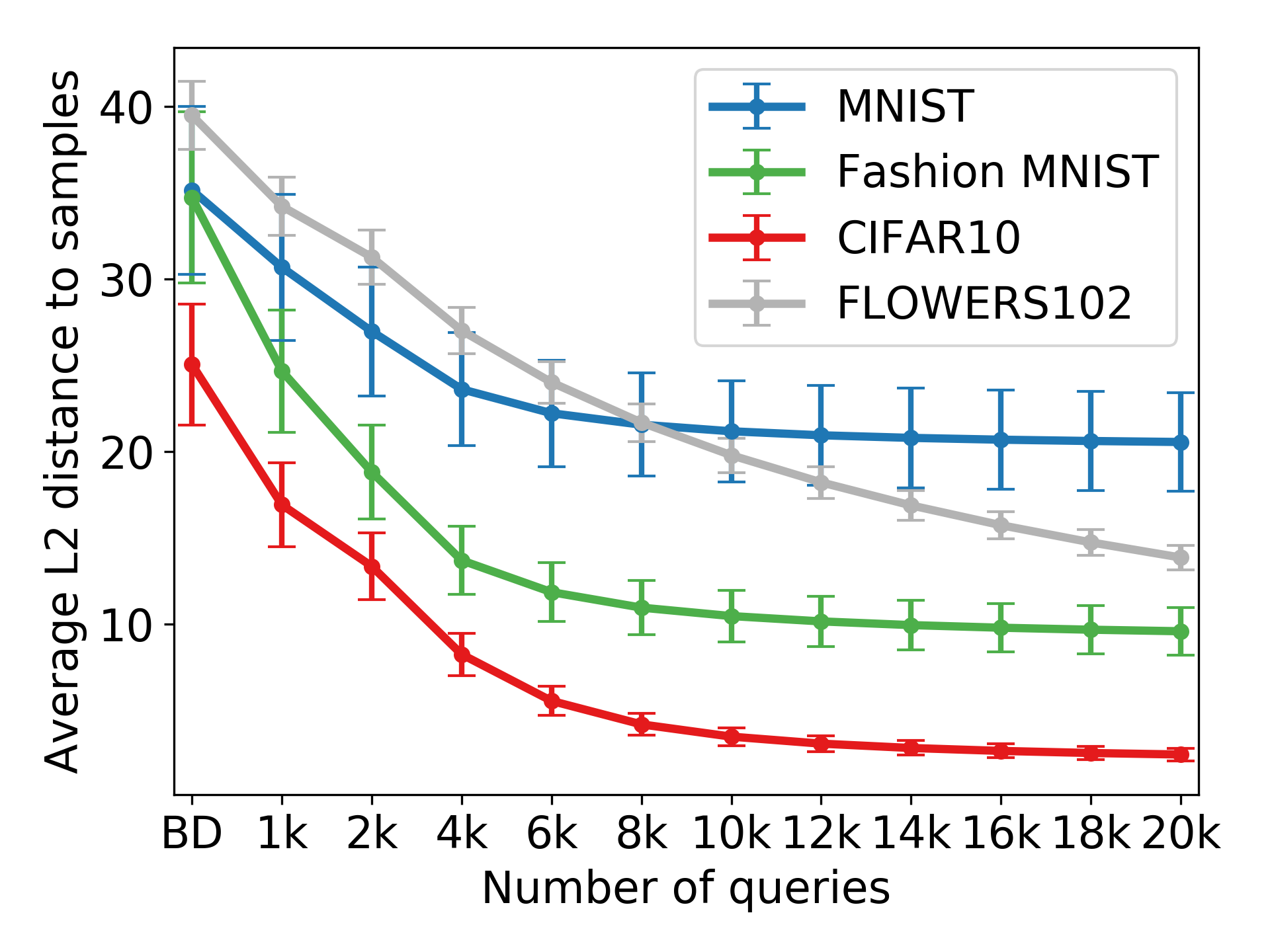}}
\subfigure[MNIST]{
\includegraphics[width=0.24\linewidth]{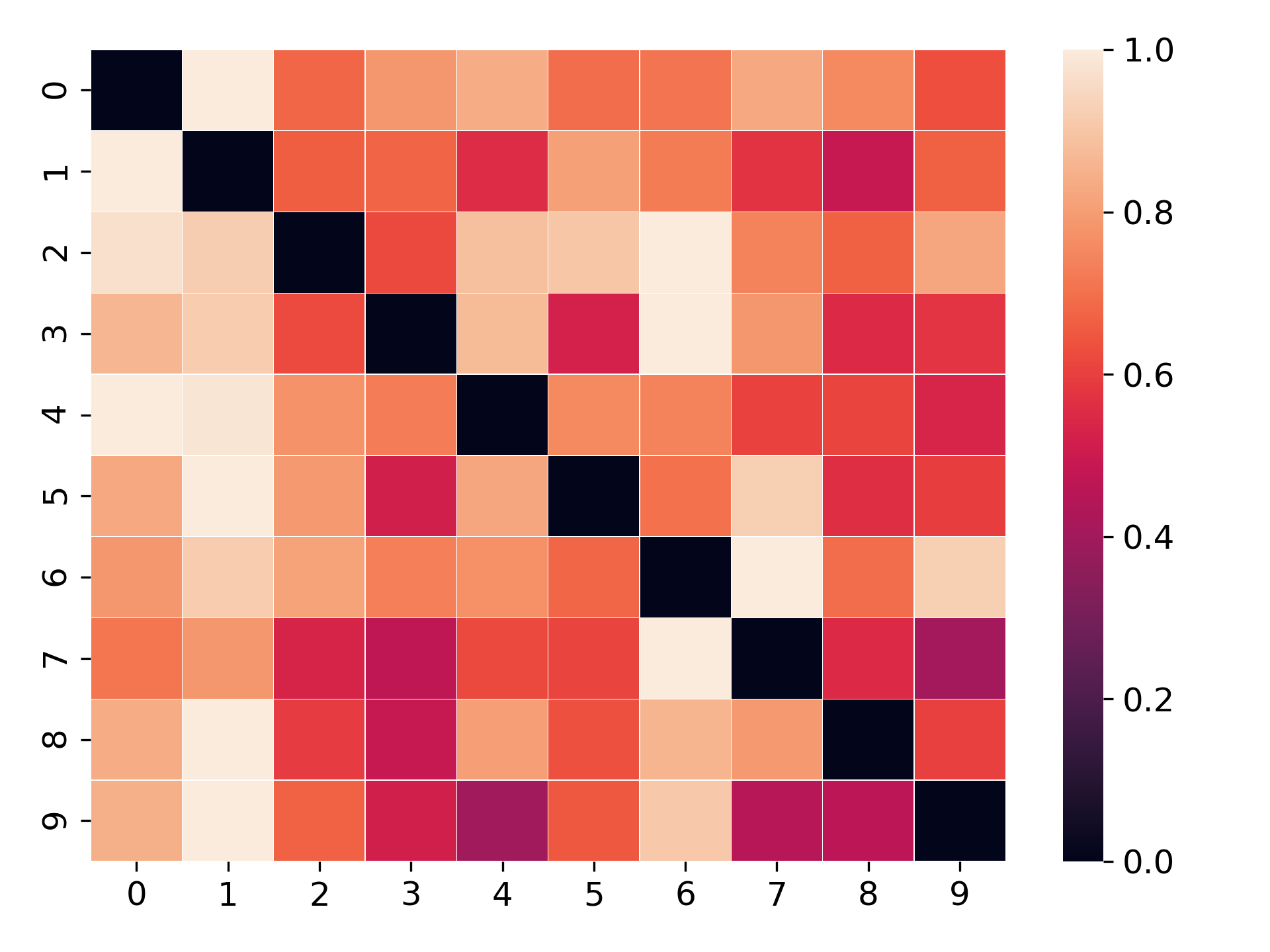}}
\subfigure[Fashion-MNIST]{
\includegraphics[width=0.24\linewidth]{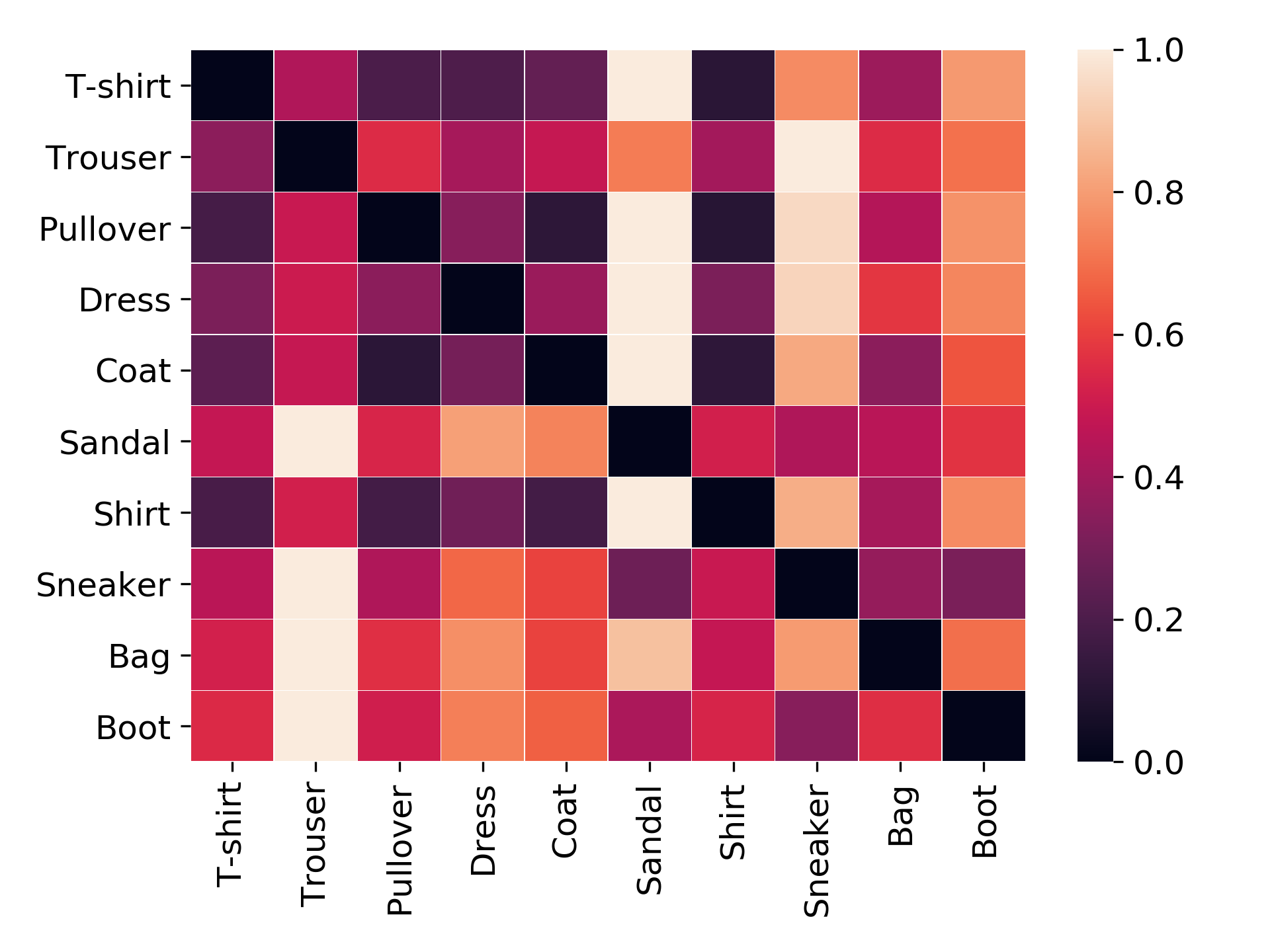}}
\subfigure[CIFAR10]{
\includegraphics[width=0.24\linewidth]{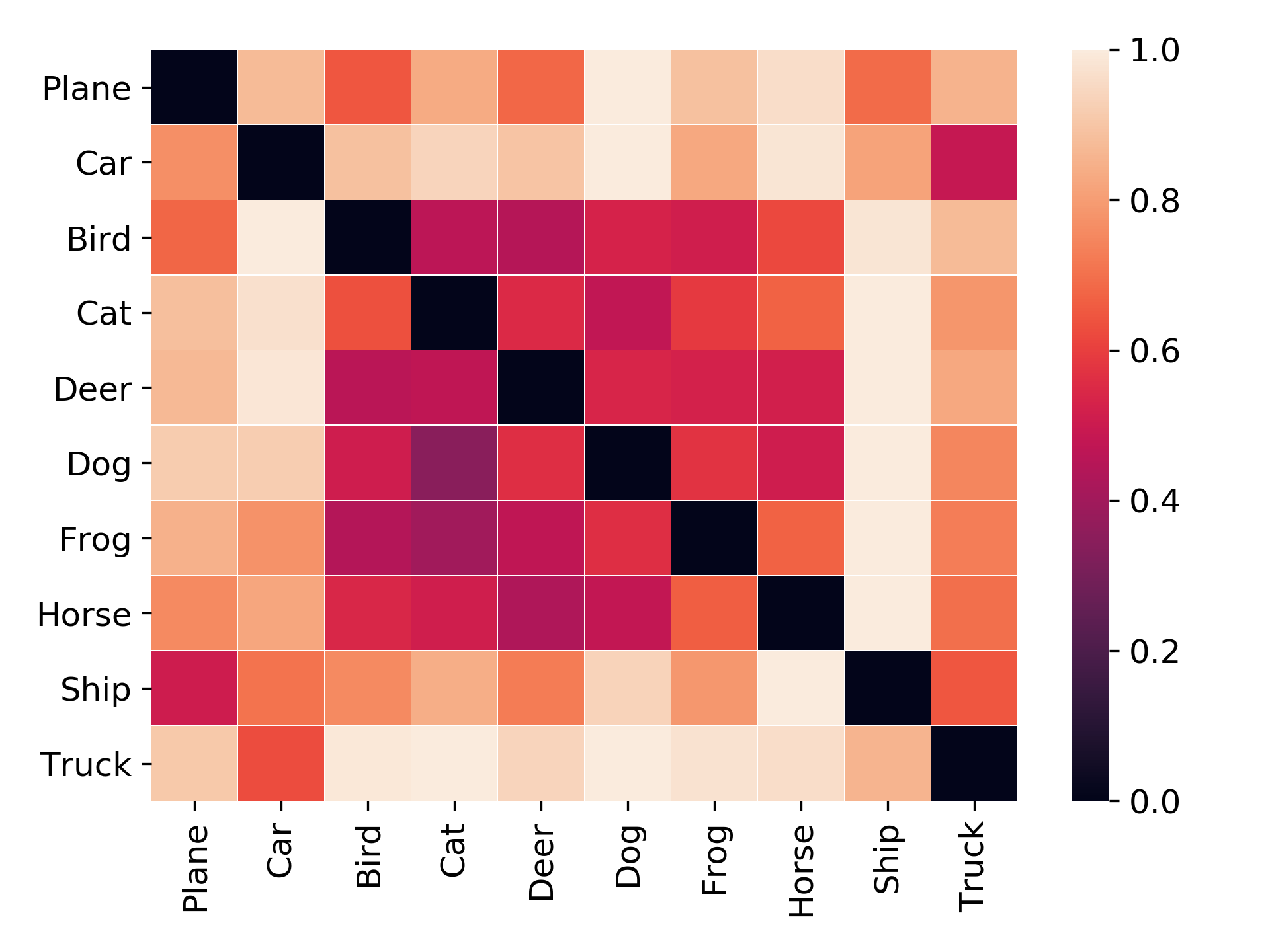}}
\end{center}
\vskip -0.13in
\caption{(a) The average minimal boundary distances over number of queries. Error bar indicates one standard deviation. (b-d) Normalized average minimal boundary distances of the samples of different classes. Darker colors indicate smaller distances between two classes.}
\label{fig:dist}
\vskip -0.13in
\end{figure*}

\subsection{Ablation Studies and Analyses of DB3KD}
We conduct several ablation studies and analyses for further understanding of the effectiveness of DB3KD.

{\noindent \bf Number of queries in label construction.} We first investigate whether different numbers of queries used for computing sample robustness has any influence on the performance. For each dataset, we query from the teacher for a variety of times from 1000 to 20000 to compute the sample robustness (Fig.~\ref{fig:iters}). It can be observed that with more queries, the student models perform slightly better, especially for deeper architectures (ResNet) and high-resolution datasets (FLOWERS102). In general, the student models perform well with various numbers of queries. Even using a binary search with around 100 queries (DB3KD-BD), the performance are satisfactory on all student models. This is because the quality of a sample's soft label is largely related to its robustness against different classes. Moreover, the MBD used for computing sample robustness shows a highly positive correlation with the number of queries (Fig.~\ref{fig:dist}(a)). The ratios of sample robustness against different classes remain stable against the number of queries. Therefore, it is not necessary to optimize the MBD with a large number of queries, which indicates that DB3KD is query efficient. It is also worth noting that the performance is not linearly correlated with the query numbers. This is because for all experiments, we use the same set of hyperparameters for fair comparison, which may not be optimal as the query number increases. However, we'd like to emphasize the performance is not sensitive to query numbers and is satisfactory with a wide range of numbers (from 2k to 20k).

Although the boundary may be complex in the pixel domain and the boundary sample may be fragile, what we actually care about is the minimal boundary distance (MBD). It actually measures how fragile a training sample is against other classes and is a robust measurement. As supplementary evidence, the standard deviations of the MDBs are relatively small (shown with the error bars in Fig.~\ref{fig:dist}(a)), indicating the robustness of the proposed approach.

\begin{table}[t]
\centering
\begin{tabular}{|C{1.83cm}|C{0.6cm}|C{1.2cm}|C{1.0cm}|C{1.2cm}|}
\hline
Algorithm & Data & Model & MNIST & FMNIST \\
\hline
\hline
Teacher CE & Yes & White & 99.33\% & 91.63\% \\
Student CE & Yes & White & 99.11\% & 90.21\% \\
Standard KD & Yes & Black-S &  99.33\% & 90.82\% \\
FSKD & Few & White &  86.70\% & 72.60\%\\
BBKD & Few & Black-S & 98.74\% & 80.90\% \\
Meta KD & Meta & White &  92.47\% & - \\
DAFL & No & White &  98.20\% & - \\
ZSKD & No & White &  98.77\% & 79.62\% \\
DFKD & No & White &  99.08\% & - \\
{\bf ZSDB3KD} & No & Black-D &  96.54\% & 72.31\% \\
\hline
\end{tabular}
\caption{Result of ZSDB3KD with MNIST and Fashion-MNIST. S: score-based teacher. D: decision-based teacher.}
\label{tab:zsDB3_mnist}
\vskip -0.2in
\end{table}

{\noindent \bf Correlation between sample robustness and class probability.}
To further analyze the effectiveness of DB3KD for constructing soft labels, we visualize the normalized average MBDs of the samples with different classes (Fig.~\ref{fig:dist}(b-d)). It is observed that classes semantically closer with each other are with smaller distances to their decision boundary. For example, in MNIST, the distance between `8' and `9' is smaller than `8' and `1' because `8' looks more like `9' than `1'. Therefore, a sample of `8' is assigned higher confidence in class `9' than `1'. Similarly, in Fashion-MNIST, `T-shirt' looks more like `shirt' than `sneaker' so that their distance are smaller. In CIFAR-10, samples of the `dog' class are with smaller distances to the boundary with `cat' than `truck' since `dog' and `cat' are semantically closer. These analyses confirm the consistency between sample robustness and class probability distribution.

\subsection{Experiment Setup of ZSDB3KD}
We evaluate ZSDB3KD with (1) a LeNet-5 and a LeNet-5-Half (on MNIST and Fashion-MNIST), and (2) an AlexNet and an AlexNet-Half (on CIFAR-10) as the teacher and the student. The networks are the same as in Section \ref{sec:setup_DB3}. 

We optimize the pseudo samples for 40 ($\xi_o=0.5$) and 100 iterations ($\xi_o=3.0$) for the two LeNet-5 and the AlexNet experiments, respectively. The query is limited to 5000 when iteratively searching for the MBD. We generate 8000 samples for each class with a batch size of 200 for all the experiments. We use data augmentation to enrich the transfer set (see Appendix). We use 5000 queries for computing the sample robustness since we have shown the number of queries is trivial. Other parameters are the same as the DB3KD experiments. We compare the performance of ZSDB3KD with several popular KD approaches in more relaxed scenarios, including FSKD \cite{kimura2018few}, BBKD \cite{wang2020neural}, Meta KD \cite{lopes2017data}, DAFL \cite{chen2019data}, ZSKD \cite{nayak2019zero} and DFKD \cite{wang2021data}. 

\begin{table}[t]
\centering
\begin{tabular}{|c|c|c|c|}
\hline
Algorithm & Data & Model & Accuracy \\
\hline
\hline
Teacher CE & Yes & White & 79.30\% \\
Student CE & Yes & White & 77.28\% \\
Standard KD & Yes & Black-S &  77.81\% \\
FSKD & Few & White &  40.58\% \\
BBKD & Few & Black-S & 74.60\% \\
DAFL & No & White &  66.38\% \\
ZSKD & No & White &  69.56\% \\
DFKD & No & White &  73.91\% \\
Noise input & No & Black-S &  14.79\% \\
Noise input & No & Black-D &  13.53\% \\
{\bf ZSDB3KD} & No & Black-D &  59.46\% \\
\hline
\end{tabular}
\caption{Result of ZSDB3KD on AlexNet with CIFAR-10.}
\label{tab:zsDB3_cifar10}
\vskip -0.15in
\end{table}
\begin{figure*}[t]
\begin{center}
\includegraphics[width=0.98\linewidth]{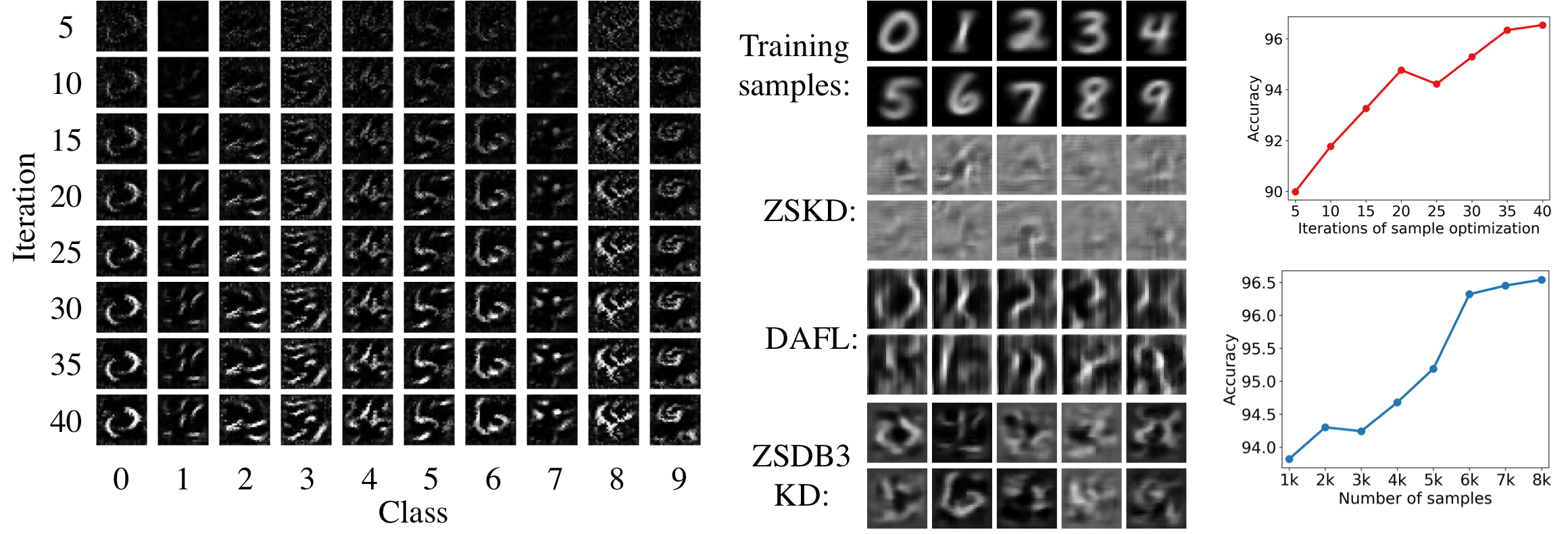}
\end{center}
\vskip -0.15in
\caption{Analysis and ablation study of ZSDB3KD with MNIST. Left: evolution of pseudo images over iterations. Middle: averaged images compared to other white-box zero-shot KD approaches. Upper right: the accuracies with different iterations of sample generation. Bottom right: the accuracies with different numbers of samples used for training the student.}
\label{fig:zsDB3_study}
\vskip -0.15in
\end{figure*}

\subsection{Performance Comparison of ZSDB3KD}
The performance of ZSDB3KD on MNIST and Fashion-MNIST, and CIFAR-10 presented in Table~\ref{tab:zsDB3_mnist} and \ref{tab:zsDB3_cifar10} show that ZSDB3KD achieves competitive performance.
The accuracies of the student networks are $96.54\%$ and $72.31\%$ on MNIST and Fashion-MNIST, which are quite close to other KD approaches with more relaxed scenarios (training data or the teacher's parameters are accessible). On CIFAR-10, our AlexNet-Half model achieves an accuracy of 59.46\% without accessing any training samples and the softmax outputs of the teacher. It is worth noting that using random noise as the input results in very poor performance with a DB3 teacher. These results indicate that the samples generated with our proposed approach indeed capture the distribution of the samples used for training the teachers. 

\subsection{Ablation Studies and Analyses of ZSDB3KD}
In this subsection, we perform several studies to understand the effectiveness of ZSDB3KD, using LeNet-5-Half trained on MNIST as an example.

{\noindent \bf Iteration of sample generation.}
We first evaluate the performance of the student with pseudo samples generated with different iterations (Fig.~\ref{fig:zsDB3_study}(upper right)). As expected, the performance is improved as the samples are optimized away from the decision boundaries with more iterations. As shown in Fig.~\ref{fig:zsDB3_study}(left), with more steps, more pixels in the pseudo samples are activated, with sharper edges and recognizable digits, which indicates that the samples become more robust as we keep moving them to the opposite of the gradient direction on the decision boundaries.

{\noindent \bf Number of samples used for training.}
We then investigate the effect of the number of pseudo samples used for training on the performance of the student network. The results of training the student network with different numbers of generated samples (from 1k to 8k per class) are presented in Fig.~\ref{fig:zsDB3_study}(bottom right). Not surprisingly, with more samples, the test accuracy increases. Even with a small number of samples (1k per class), the student network can still achieve a competitive performance of 94\% test accuracy. With 8k samples per class, the student's performance gets saturated and is comparable to the performance of standard KD.

{\noindent \bf Visualization of generated samples.}
As mentioned above, we have shown the evolution of individual samples over iterations (Fig.~\ref{fig:zsDB3_study}(left)), which gradually exhibits clear digits. To have a further visualization of the generated pseudo samples, we further average 1k samples for each class as shown in Fig.~\ref{fig:zsDB3_study}(middle). Even though generated with a DB3 teacher, the samples are with a satisfactory quality compared with the averaged samples generated with ZSKD and DAFL that use white-box teachers. 


\section{Conclusion}
In this study, we introduced KD from a decision-based black-box teacher for the first time. We proposed DB3KD to deal with this problem, which uses sample robustness to construct the soft labels for the training samples by iteratively querying from the teacher. We also extend DB3KD to a much more challenging scenario in which the training set is not accessible and named it Zero-shot DB3KD (ZSDB3KD). Experiments on various networks and datasets validated the effectiveness of the proposed approaches.

Our study motivated a new line of research on KD, in which the black-box teacher only returns top-1 classes.  It is a much more challenging scenario because the class probabilities of the training samples need to be constructed by iteratively querying from the DB3 teacher. With the training set accessible, our DB3KD achieved competitive performance on FLOWERS102, in which samples largely overlap with ImageNet. We believe that DB3KD can work effectively on large-scale datasets. With the training samples not available, like most of the existing works, a large amount of computing resource is required for pseudo sample generation, making zero-shot KD hard to accomplish with large-scale datasets. With a DB3 teacher, even more iterations are needed compared to learning from a white-box model. Although we proposed the first principled solution, we hope it helps to raise attention in this area and promote efficient approaches.

\bibliography{example_paper}
\bibliographystyle{icml2021}

\clearpage
\onecolumn
\setcounter{table}{0}
\renewcommand{\thetable}{S\arabic{table}}
\setcounter{figure}{0}
\renewcommand{\thefigure}{S\arabic{figure}}
\section*{Appendix}
\subsection*{A. Architectures Used in DB3KD and ZSDB3KD Experiments}
We use several networks to evaluate the performance of DB3KD and ZSDB3KD. LeNet-5 (teacher)/ LeNet-5-Half (student) and AlexNet (teacher)/ AlexNet-Half (student) are used for both DB3KD and ZSDB3KD experiments (Table~\ref{tab:lenet5}-\ref{tab:alexnet-half}). For the DB3KD experiments, we also design two student networks, i.e., LeNet-5-1/5 and AlexNet-Quarter for further evaluation (Table~\ref{tab:lenet5-1/5}-\ref{tab:alexnet-quarter}). We also conduct experiments with ResNet-34 (teacher)/ ResNet-18 (student) with DB3KD and the architectures used are the same as the original ResNet architectures.

\begin{table}[htb]
\centering
\begin{tabular}{cccccccc}
\hline
Index & Layer & Type & Feature map & Kernel size & Stride & Padding & Activation \\
0 & Input & Input & 1 & - & - & - & - \\
1 & conv1 & conv & 20 & 5x5 & 1 & 0 & ReLu \\
2 & maxpool1 & pooling & - & 2x2 & 2 & 1 & - \\
3 & conv2 & conv & 50 & 5x5 & 1 & 0 & ReLu \\
4 & maxpool2 & pooling & - & 2x2 & 2 & 1 & - \\
5 & fc1 & fc & 200 & - & - & - &  ReLu\\
6 & fc2 & fc & 10 & - & - & - &  Softmax\\
\hline
\end{tabular}
\vskip -0.1in
\caption{The architecture of the teacher model LeNet-5 with MNIST and Fashion-MNIST, for both DB3KD and ZSDB3KD experiments.}
\label{tab:lenet5}
\vskip -0.2in
\end{table}

\begin{table}[htb]
\centering
\begin{tabular}{cccccccc}
\hline
Index & Layer & Type & Feature map & Kernel size & Stride & Padding & Activation \\
0 & Input & Input & 1 & - & - & - & - \\
1 & conv1 & conv & 10 & 5x5 & 1 & 0 & ReLu \\
2 & maxpool1 & pooling & - & 2x2 & 2 & 1 & - \\
3 & conv2 & conv & 25 & 5x5 & 1 & 0 & ReLu \\
4 & maxpool2 & pooling & - & 2x2 & 2 & 1 & - \\
5 & fc1 & fc & 100 & - & - & - &  ReLu\\
6 & fc2 & fc & 10 & - & - & - &  Softmax\\
\hline
\end{tabular}
\vskip -0.1in
\caption{The architecture of the student model LeNet-5-Half with MNIST and Fashion-MNIST, for both DB3KD and ZSDB3KD experiments.}
\label{tab:lenet5-half}
\vskip -0.2in
\end{table}

\begin{table}[htb]
\centering
\begin{tabular}{ccccccccc}
\hline
Index & Layer & Type & Feature map & Kernel size & Stride & Padding & Activation \\
0 & Input & Input & 1 & - & - & - &  - \\
1 & conv1 & conv & 64 & 3x3 & 2 & 1 &  ReLu \\
2 & maxpool1 & pooling & - & 3x3 & 2 & 0 &  - \\
3 & bn1 & batch norm  & - & - & - & - &  - \\
4 & conv2 & conv & 192 & 3x3 & 1 & 2 &  ReLu \\
5 & maxpool2 & pooling & - & 3x3 & 2 & 0 & - \\
6 & bn2 & batch norm  & - & - & - & - &  - \\
7 & conv3 & conv & 384 & 3x3 & 1 & 1 &  ReLu \\
8 & bn3 & batch norm  & - & - & - & - &  - \\
9 & conv4 & conv & 256 & 3x3 & 1 & 1 &  ReLu \\
10 & bn4 & batch norm  & - & - & - & - & - \\
11 & conv5 & conv & 256 & 3x3 & 1 & 1 & ReLu \\
12 & maxpool3 & pooling & - & 3x3 & 2 & 0 & - \\
13 & bn5 & batch norm  & - & - & - & - & - \\
14 & fc1 & fc & 4096 & - & - & - & ReLu\\
15 & bn6 & batch norm  & - & - & - & - & - \\
16 & fc2 & fc & 4096 & - & - & - & ReLu\\
17 & bn7 & batch norm  & - & - & - & - & - \\
18 & fc3 & fc & 10 & - & - & - & Softmax\\
\hline
\end{tabular}
\caption{The architecture of the teacher model AlexNet with CIFAR-10, for both DB3KD and ZSDB3KD experiments.}
\label{tab:alexnet}
\end{table}

\begin{table}[htb]
\centering
\begin{tabular}{ccccccccc}
\hline
Index & Layer & Type & Feature map & Kernel size & Stride & Padding & Activation \\
0 & Input & Input & 1 & - & - & - &  - \\
1 & conv1 & conv & 32 & 3x3 & 2 & 1 &  ReLu \\
2 & maxpool1 & pooling & - & 3x3 & 2 & 0 &  - \\
3 & bn1 & batch norm  & - & - & - & - &  - \\
4 & conv2 & conv & 96 & 3x3 & 1 & 2 &  ReLu \\
5 & maxpool2 & pooling & - & 3x3 & 2 & 0 & - \\
6 & bn2 & batch norm  & - & - & - & - &  - \\
7 & conv3 & conv & 192 & 3x3 & 1 & 1 &  ReLu \\
8 & bn3 & batch norm  & - & - & - & - &  - \\
9 & conv4 & conv & 128 & 3x3 & 1 & 1 &  ReLu \\
10 & bn4 & batch norm  & - & - & - & - & - \\
11 & conv5 & conv & 128 & 3x3 & 1 & 1 & ReLu \\
12 & maxpool3 & pooling & - & 3x3 & 2 & 0 & - \\
13 & bn5 & batch norm  & - & - & - & - & - \\
14 & fc1 & fc & 2048 & - & - & - & ReLu\\
15 & bn6 & batch norm  & - & - & - & - & - \\
16 & fc2 & fc & 2048 & - & - & - & ReLu\\
17 & bn7 & batch norm  & - & - & - & - & - \\
18 & fc3 & fc & 10 & - & - & - & Softmax\\
\hline
\end{tabular}
\vskip -0.1in
\caption{The architecture of the student model AlexNet-Half with CIFAR-10, for both DB3KD and ZSDB3KD experiments.}
\label{tab:alexnet-half}
\vskip -0.2in
\end{table}

\begin{table}[htb]
\centering
\begin{tabular}{cccccccc}
\hline
Index & Layer & Type & Feature map & Kernel size & Stride & Padding & Activation \\
0 & Input & Input & 1 & - & - & - & - \\
1 & conv1 & conv & 4 & 5x5 & 1 & 0 & ReLu \\
2 & maxpool1 & pooling & - & 2x2 & 2 & 1 & - \\
3 & conv2 & conv & 10 & 5x5 & 1 & 0 & ReLu \\
4 & maxpool2 & pooling & - & 2x2 & 2 & 1 & - \\
5 & fc1 & fc & 40 & - & - & - &  ReLu\\
6 & fc2 & fc & 10 & - & - & - &  Softmax\\
\hline
\end{tabular}
\vskip -0.1in
\caption{The architecture of the student model LeNet-5-1/5 with MNIST and Fashion-MNIST, for DB3KD experiments.}
\label{tab:lenet5-1/5}
\vskip -0.2in
\end{table}

\begin{table}[htb]
\centering
\begin{tabular}{ccccccccc}
\hline
Index & Layer & Type & Feature map & Kernel size & Stride & Padding & Activation \\
0 & Input & Input & 1 & - & - & - &  - \\
1 & conv1 & conv & 16 & 3x3 & 2 & 1 &  ReLu \\
2 & maxpool1 & pooling & - & 3x3 & 2 & 0 &  - \\
3 & bn1 & batch norm  & - & - & - & - &  - \\
4 & conv2 & conv & 48 & 3x3 & 1 & 2 &  ReLu \\
5 & maxpool2 & pooling & - & 3x3 & 2 & 0 & - \\
6 & bn2 & batch norm  & - & - & - & - &  - \\
7 & conv3 & conv & 96 & 3x3 & 1 & 1 &  ReLu \\
8 & bn3 & batch norm  & - & - & - & - &  - \\
9 & conv4 & conv & 64 & 3x3 & 1 & 1 &  ReLu \\
10 & bn4 & batch norm  & - & - & - & - & - \\
11 & conv5 & conv & 64 & 3x3 & 1 & 1 & ReLu \\
12 & maxpool3 & pooling & - & 3x3 & 2 & 0 & - \\
13 & bn5 & batch norm  & - & - & - & - & - \\
14 & fc1 & fc & 1024 & - & - & - & ReLu\\
15 & bn6 & batch norm  & - & - & - & - & - \\
16 & fc2 & fc & 1024 & - & - & - & ReLu\\
17 & bn7 & batch norm  & - & - & - & - & - \\
18 & fc3 & fc & 10 & - & - & - & Softmax\\
\hline
\end{tabular}
\caption{The architecture of the student model AlexNet-Quarter with CIFAR-10, for DB3KD experiments.}
\label{tab:alexnet-quarter}
\end{table}
 
\clearpage
\subsection*{B. Experiment details}
\subsubsection*{\bf B.1. Training of the Models with Cross-Entropy}
In this subsection, we introduce the details of training the models with cross-entropy loss, for both the pre-trained models used as the DB3 teachers, and the performance of the student models trained solely with the cross-entropy loss reported in Tables \ref{tab:main}, \ref{tab:selfkd}, \ref{tab:zsDB3_mnist}, and \ref{tab:zsDB3_cifar10}.

{\bf LeNet-5 on MNIST and Fashion-MNIST}
For the LeNet-5 architecture on MNIST and Fashion-MNIST, we train the teacher model for 200 epochs, with a batch size of 1024, an Adam optimizer with a learning rate of 0.001. For the student models trained with cross-entropy (reported in Tables \ref{tab:main} and \ref{tab:zsDB3_mnist}), we use the same hyperparameters as above.

{\bf AlexNet on CIFAR-10}
For the AlexNet architecture on CIFAR-10, we train the teacher model for 300 epochs, with a batch size of 1024 and an SGD optimizer. We set the momentum to 0.9, and weight decay to 0.0001. The learning rate is set to 0.1 at the beginning, and is divided by 10 at epochs 60, 120, and 180. For the student models trained with cross-entropy (reported in Tables \ref{tab:main} and \ref{tab:zsDB3_cifar10}), we use the same hyperparameters as above.

{\bf ResNet on CIFAR-100} For the ResNet-\{50,34\} on CIFAR-100, we train the teacher models for 300 epochs, with a batch size of 256 and an SGD optimizer. We set the momentum to 0.9 and weight decay to 0.0001. The learning rate is set to 0.1 at the beginning, and is divided by 10 at epochs 60, 120, and 180. For the student model (ResNet-18) trained with cross-entropy (reported in Table \ref{tab:selfkd}), we use the same hyperparameters as above.

{\bf ResNet-34 on FLOWERS102}
For the ResNet-34 architecture on FLOWERS102, we start with the model pre-trained on ImageNet, which is provided by Pytorch, and fine-tune the pre-trained model for 200 epochs with an SGD optimizer. We set the batch size to 64 and the momentum to 0.9. The learning rate is set to 0.01 at the beginning, and set to 0.005 and 0.001 at epochs 60 and 100, respectively. For the student model (Resnet-18) trained with cross-entropy (reported in Table \ref{tab:main}), we use the same hyperparameters as above.

\subsubsection*{\bf B.2. Standard Knowledge Distillation Training Details}
For the standard knowledge distillation results reported in Tables \ref{tab:main}, \ref{tab:selfkd}, \ref{tab:zsDB3_mnist}, and \ref{tab:zsDB3_cifar10}, we train the student models via standard KD with the following hyperparameters. The scaling factor $\lambda$ that balances the importance of cross-entropy loss and knowledge distillation loss is set to 1. The Adam optimizer is used for all experiments and the student networks are trained for 200 epochs with a temperature of 20. For the experiments with MNIST, Fashion-MNIST, and CIFAR-10, we set the batch size to 512; for the experiments with CIFAR-100 and FLOWERS102, we set the batch size to 64. The learning rate is set to 0.001 for MNIST and Fashion-MNIST, 0.005 for CIFAR-10/100, and 0.0005 for FLOWERS102.

\subsubsection*{\bf B.3. Surrogate Knowledge Distillation Training Details}
Training the student networks by transferring the knowledge from a surrogate, low-capacity white-box teacher whose parameters can be fully accessed is sensitive to hyperparameter selection. We did an extensive hyperparameter search in our experiments and report the best numbers in Table \ref{tab:main}. We use the hyperparameters listed below. The optimizer and batch size used for surrogate KD are the same as in standard KD. We train the student models for 300 epochs for all experiments. For MNIST and Fashion-MNIST, the scaling factor $\lambda$ is set to 0.7, the temperature is set to 3, and the learning rate is set to 0.005. For CIFAR-10/100, $\lambda$ is set to 0.5, the temperature is set to 5, and the learning rate is set to 0.005. For FLOWERS102, $\lambda$ is set to 1, the temperature is set to 10, and the learning rate is set to 0.001. 

\subsubsection*{\bf B.4. Data Augmentation Used in ZSDB3KD Experiments}
In ZSDB3KD experiments, we found that data augmentation can improve the performance. Since the number of queries for the soft label construction of the samples is trivial to the performance, as shown in the DB3KD experiments (Fig. \ref{fig:iters}), we can apply various augmentation strategies to enrich the transfer set with affordable extra computing cost. In our study, we implement the following data augmentation strategies.

\begin{itemize}
    \item {\bf (1) Padding and crop.} We first pad two pixels on each side of the generated samples and crop it to the original size, starting from the upper left corner to the bottom right corner, with an interval of 1.
    \item {\bf (2) Horizontal and vertical flip.} We flip the generated samples horizontally and vertically to create mirrored samples.
    \item {\bf (3) Rotation.} We rotate each generated image starting from $-15^{\circ}$ to $15^{\circ}$ with an interval of $5^{\circ}$ to create 6 more rotated samples.
    \item {\bf (4) Flip after padding and crop.} We flip the images after (1), horizontally and vertically.
    \item {\bf (5) Rotation after padding and crop.} We rotate the images after (1), using the same operation as (3).
\end{itemize}

For the MNIST and Fashion-MNIST datasets, only the strategies (1) and (2) are used. For the CIFAR-10 dataset, all five strategies are used. For the DB3KD experiment with CIFAR-100, we also use the above five strategies.

It is also worth mentioning that after generating images with the above operations, some of the samples' top-1 classes change to others. If this happens, we use the approach described in Section 3 to find the sample's corresponding point on the targeted decision boundary, i.e., $x^*$, to recover its top-1 class back to the top-1 class of the sample before augmentation.

Table~\ref{tab:aug} presents the performance comparison with and without data augmentation on each dataset used in the ZSDB3KD experiments. It is observed that training the student networks with more samples augmented with the above strategies can improve the performance. 
\begin{table}[htb]
\centering
\begin{tabular}{ccc}
\hline
Dataset & Acc. without augmentation & Acc. with augmentation \\
MNIST & 94.20\% & 96.54\% \\
Fashion-MNIST & 67.24\% & 72.31\% \\
CIFAR-10 & 37.58\% & 59.46\% \\
\hline
\end{tabular}
\caption{Performance comparison of the ZSDB3KD experiments with and without data augmentation, with LeNet-5-Half on the MNIST and Fashion-MNIST datasets, and with AlexNet-Half on the CIFAR-10 dataset, respectively.}
\label{tab:aug}
\end{table}

\subsection*{C. More Experiment Results}
\subsubsection*{\bf C.1. Comparison of the Sample Robustness Computed with DB3KD and the Logits Generated by the Teacher}
\begin{figure*}[htb]
\begin{center}
\subfigure[MNIST]{
\includegraphics[width=0.32\linewidth]{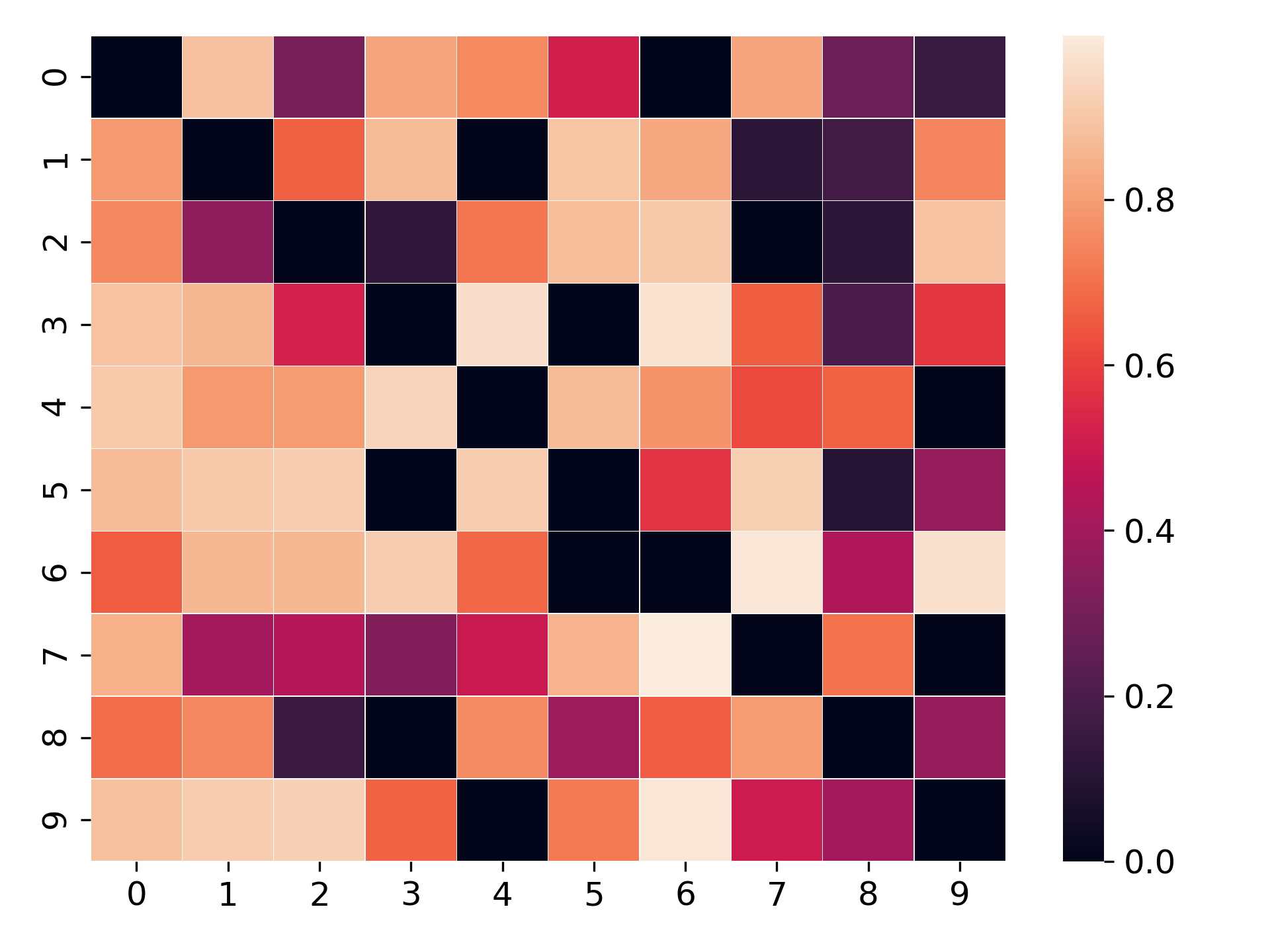}}
\subfigure[Fashion-MNIST]{
\includegraphics[width=0.32\linewidth]{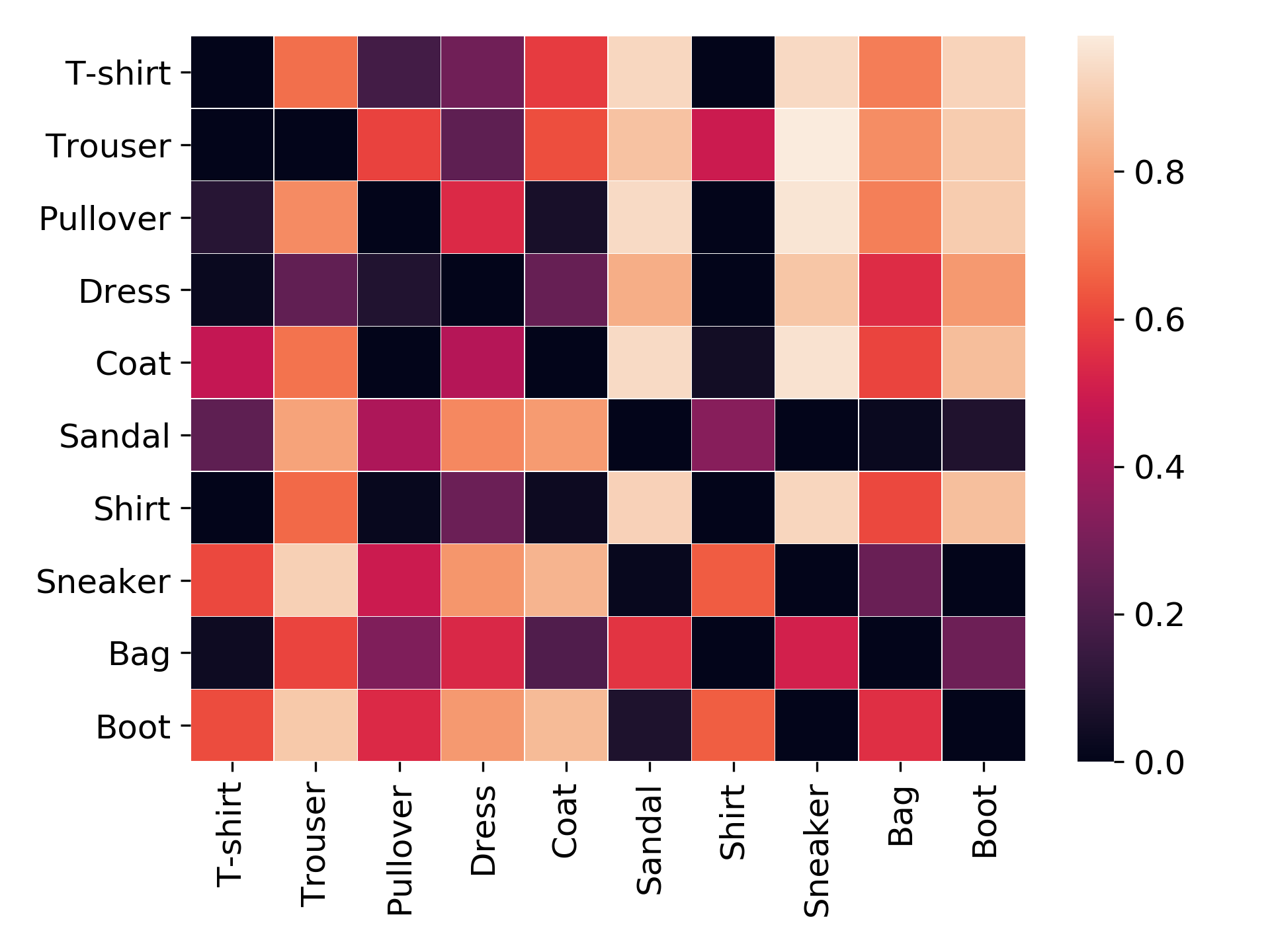}}
\subfigure[CIFAR-10]{
\includegraphics[width=0.32\linewidth]{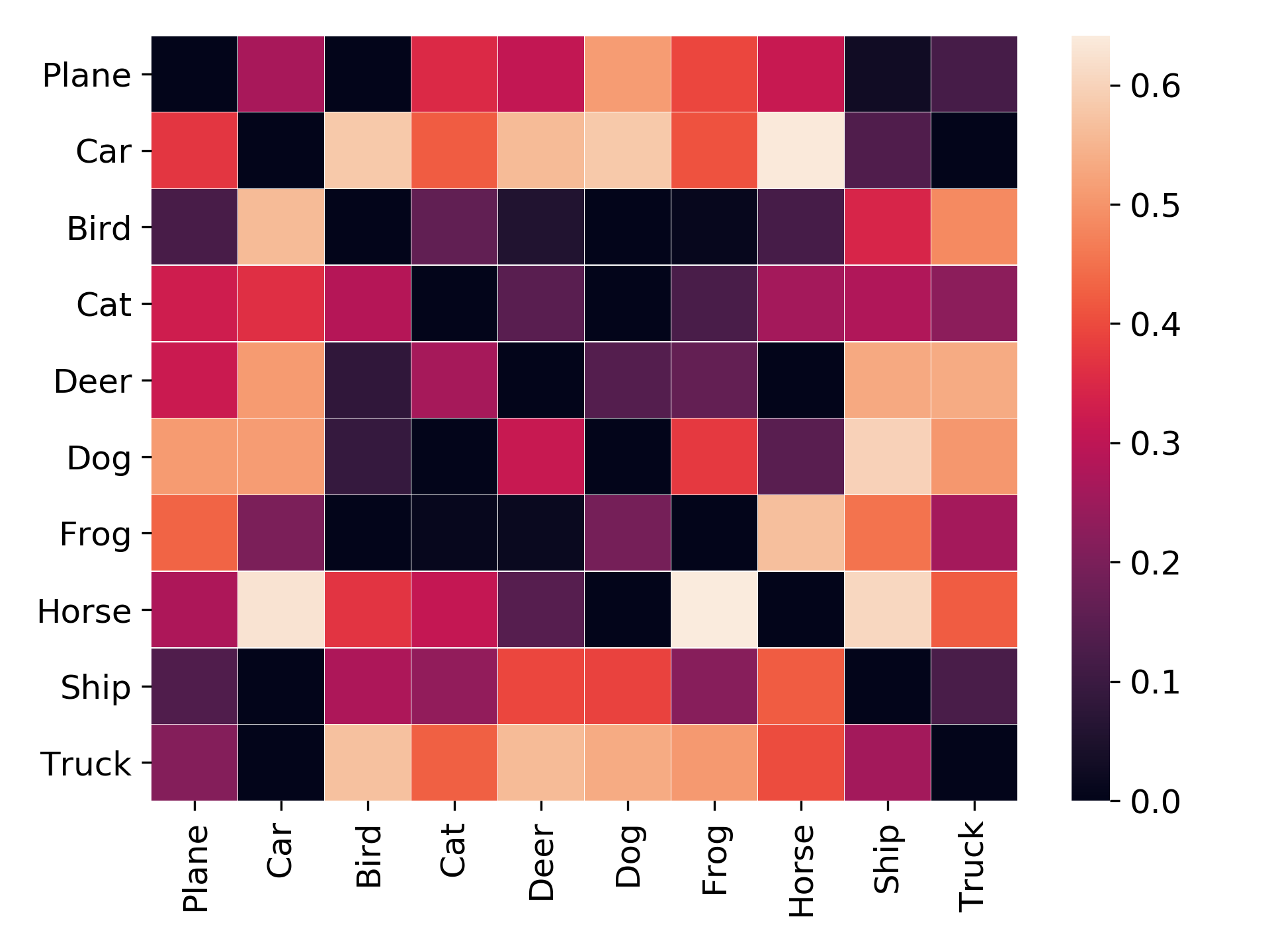}}
\end{center}
\caption{Normalized average distances of the samples of different classes, computed with the softmax outputs of the pre-trained teachers. Darker colors indicate smaller distances between two classes.}
\label{fig:dist_a}
\end{figure*}

To further understand the effectiveness of the label construction with sample robustness in our DB3KD approach, we visualize the sample distances that are computed with the softmax outputs of the teacher networks, by accessing the teachers' parameters. We first feed the training samples to the teacher model and get the softmax output. For a training sample, if a bigger probability is assigned to a class, it means the distance between this sample to the specific class is smaller. Therefore, we simply use \emph{1 - class probability} to represent the sample distance. The results are presented in Fig.~\ref{fig:dist_a}. It can be observed that the visualized heatmaps look similar to those visualized with the sample robustness computed with our DB3 approach (Fig. \ref{fig:dist}(b-d)). For example, both of the MNIST heatmaps indicate that digit '4' is close to digit '9'. For the Fashion-MNIST, Fig.~\ref{fig:dist_a}(b) shows that class T-shirt is semantically close to class 'Shirt' and 'Pullover', which is consistent with the results in Fig. \ref{fig:dist}(c). These results further validate that our proposed approach to construct soft labels with sample robustness is meaningful.

\subsubsection*{\bf C.2. Ablation Studies of ZSDB3KD on Fashion-MNIST and CIFAR-10}
\begin{figure*}[htb]
\begin{center}
\subfigure[]{
\includegraphics[width=0.35\linewidth]{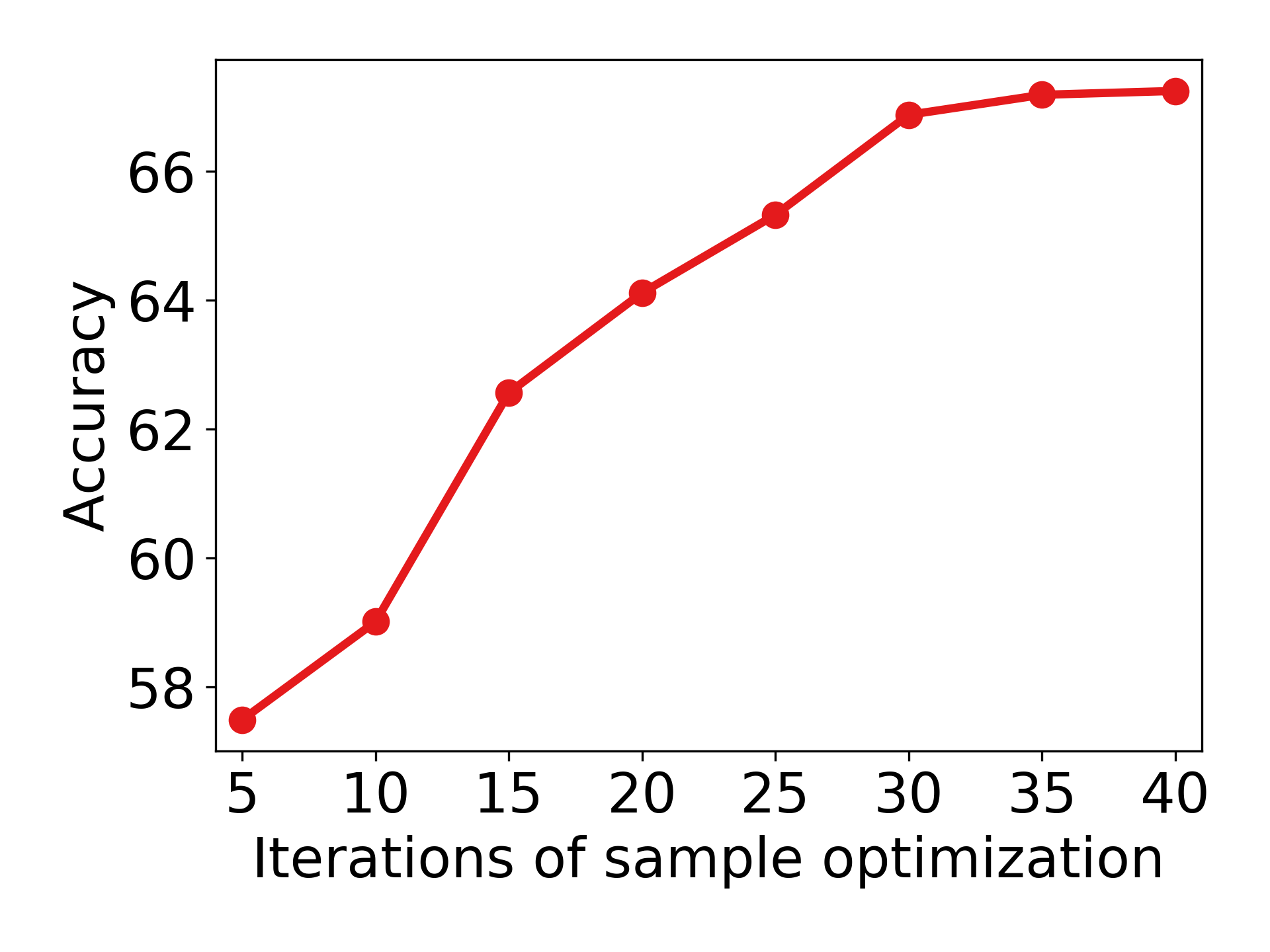}}
\subfigure[]{
\includegraphics[width=0.35\linewidth]{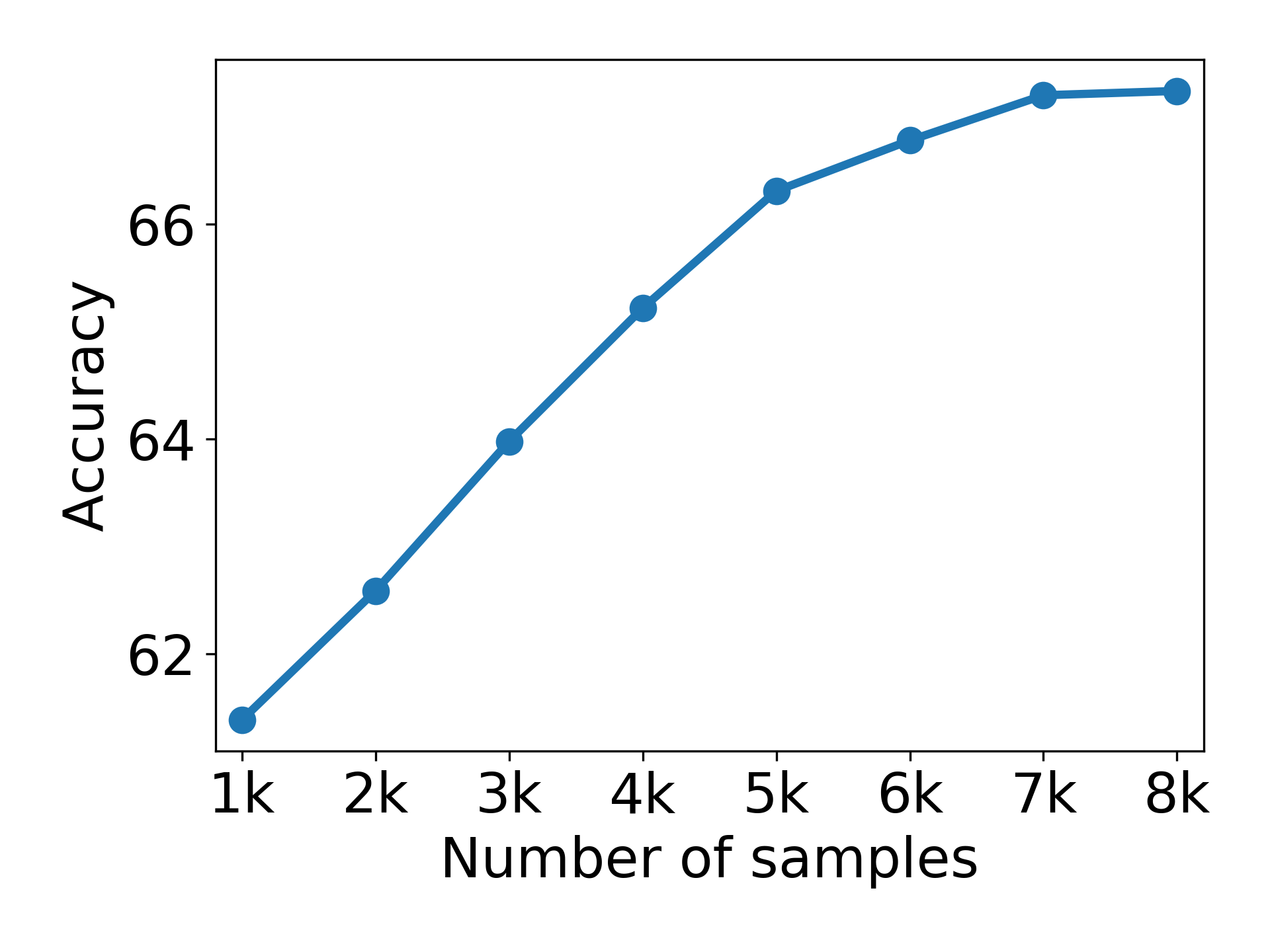}}
\end{center}
\caption{Performance of ZSDB3KD on the Fashion-MNIST dataset with (a) different numbers of iterations for sample generation and (b) pseudo samples used for KD training. Data augmentation is not used for the study.}
\label{fig:fashion}
\end{figure*}

\begin{figure*}[htb]
\begin{center}
\subfigure[]{
\includegraphics[width=0.35\linewidth]{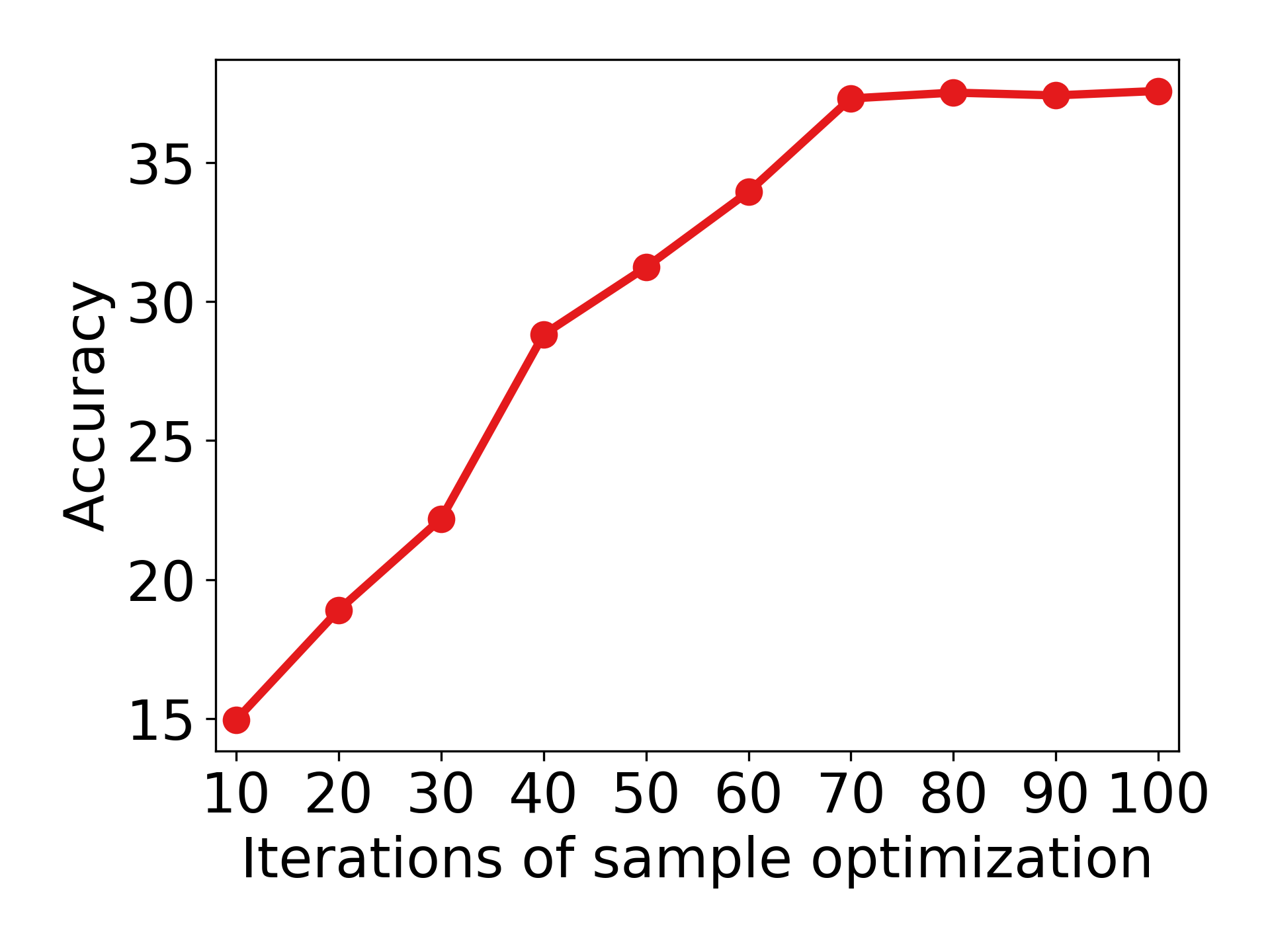}}
\subfigure[]{
\includegraphics[width=0.35\linewidth]{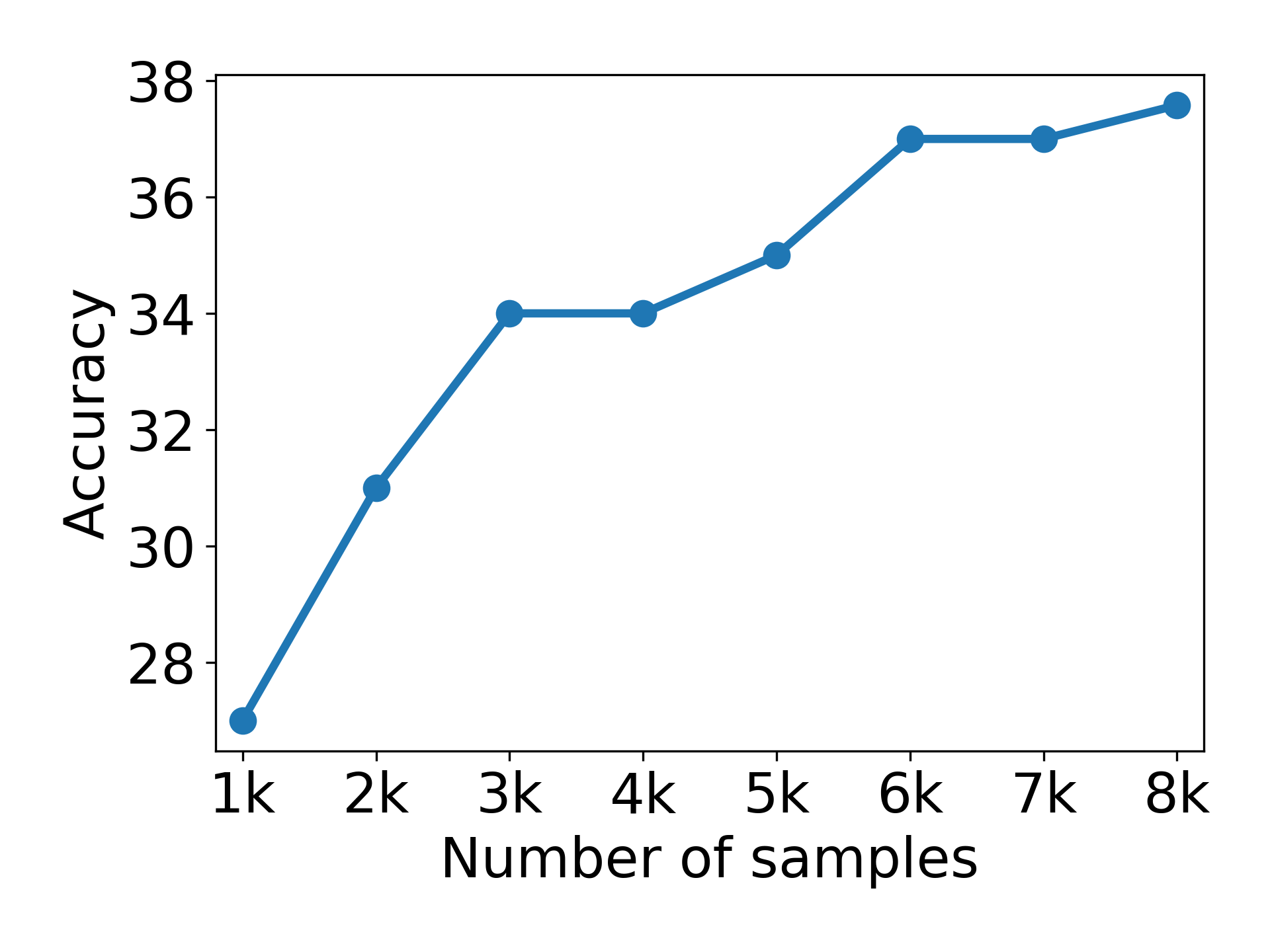}}
\end{center}
\caption{Performance of ZSDB3KD on the CIFAR-10 dataset with (a) different numbers of iterations for sample generation and (b) pseudo samples used for KD training. Data augmentation is not used for the study.}
\label{fig:cifar}
\end{figure*}

Similar to the ablation studies of ZSDB3KD on the MNIST dataset, we also investigate the effect of (1) different numbers of iterations for sample generation and (2) different numbers of pseudo samples used for KD training  on the performance of the student networks (without using data augmentation). The results are presented in Fig.~\ref{fig:fashion} and Fig.~\ref{fig:cifar}, respectively.

Similar to the results of MNIST, it is observed that, with more iterations for the sample optimization, more robust pseudo samples can be generated and the performance of the student networks are increased via DB3KD. For example, when optimizing the randomly generated noises for only 5 iterations, the performance of the student network on the Fashion-MNIST is less than 58\% without data augmentation. After 40 iterations, the performance increases by around 7\%. The performance of the AlexNet-Half network on CIFAR-10 is only around 15\% when using pseudo samples that are optimized for only 10 iterations. On the other hand, the performance increases to 37\% after 70 iterations.

The test accuracies of the student networks are also higher when using more pseudo samples as the transfer set. For the Fashion-MNIST dataset, the performance increases from 61.39\% to 67.24\% as the number of pseudo samples used as the transfer set increases from 1000 to 8000 per category. For the CIFAR-10 dataset, the performance is less than 28\% when using only 1000 samples per class. When the number of samples for each class increases to 8000, an accuracy of 37.58\% can be achieved.

\end{document}